%% file: iclr2026_conference.tex
\documentclass{article} 
\usepackage{iclr2026_conference, times}

\input{math_commands.tex}

\usepackage{hyperref}
\usepackage{url}
\usepackage[percent]{overpic}
\usepackage{adjustbox}
\usepackage{booktabs}
\usepackage{multirow}
\usepackage{array}
\newcolumntype{C}{>{\centering\arraybackslash}p{1.2cm}}
\usepackage{xcolor}
\usepackage{colortbl}
\usepackage{makecell}
\usepackage{listings}
\usepackage{wrapfig}
\usepackage{fancyvrb}
\usepackage{textcomp}

\newcommand{\name}{SpectrumWorld}
\newcommand{\lab}{SpectrumLab}
\newcommand{\bench}{SpectrumBench}
\newcommand{\omnigen}{SpectrumAnnotator}

\newcommand{\eqcontrib}{\textsuperscript{*}}
\newcommand{\corr}{\textsuperscript{\textdagger}}

\usepackage{tcolorbox} 
\tcbuselibrary{skins}

\newtcolorbox{promptboxblue}[1]{
    enhanced,
    attach boxed title to top left={xshift=5mm, yshift=-\tcboxedtitleheight/2},
    colback=blue!5,
    colframe=blue!75!black,
    colbacktitle=blue!75!black,
    coltitle=white,
    fonttitle=\bfseries,
    boxrule=1pt,
    title=#1,
    sharp corners,
    boxed title style={
        sharp corners,
        frame code={
            \path[fill=tcbcolback,draw=tcbcolframe]
            (frame.north west) -- (frame.north east) --
            (frame.south east) -- (frame.south west) -- cycle;
        }
    }
}

\title{\name: Artificial Intelligence \\
Foundation for Spectroscopy}

\author{
\textbf{Zhuo Yang}\textsuperscript{1,2}\eqcontrib \quad
\textbf{Jiaqing Xie}\textsuperscript{1}\eqcontrib \quad
\textbf{Shuaike Shen}\textsuperscript{10} \quad
\textbf{Daolang Wang}\textsuperscript{6} \quad
\textbf{Yeyun Chen}\textsuperscript{8,9} \\
\textbf{Ben Gao}\textsuperscript{1,5} \quad
\textbf{Shuzhou Sun}\textsuperscript{1,7} \quad
\textbf{Biqing Qi}\textsuperscript{1} \quad
\textbf{Dongzhan Zhou}\textsuperscript{1} \quad
\textbf{Lei Bai}\textsuperscript{1} \quad
\textbf{Linjiang Chen}\textsuperscript{4} \\
\textbf{Shufei Zhang}\textsuperscript{1} \quad
\textbf{Qinying Gu}\textsuperscript{1} \quad
\textbf{Jun Jiang}\textsuperscript{4}\corr \quad
\textbf{Tianfan Fu}\textsuperscript{3,1}\corr \quad
\textbf{Yuqiang Li}\textsuperscript{1}\corr \\
\\ 
\textsuperscript{1}Shanghai Artificial Intelligence Laboratory \quad
\textsuperscript{2}Xidian University \quad
\textsuperscript{3}Nanjing University \\
\textsuperscript{4}University of Science and Technology of China \quad
\textsuperscript{5}Wuhan University \\
\textsuperscript{6}North University of China \quad
\textsuperscript{7}Center for Machine Vision and Signal Analysis (CMVS), University of Oulu \\
\textsuperscript{8}Institute of Artificial Intelligence, Xiamen University \quad
\textsuperscript{9}Shanghai Innovation Institute \\
\textsuperscript{10}Carnegie Mellon University \\
\\
\eqcontrib\ Equal contribution \quad
\corr\ Corresponding author
}

\iclrfinalcopy 
\begin{document}
  \maketitle

  \begin{abstract}
    Deep learning holds immense promise for spectroscopy, yet research and
    evaluation in this emerging field often lack standardized formulations. To address
    this issue, we introduce \lab, a pioneering unified platform designed to systematize
    and accelerate deep learning research in spectroscopy. \lab\ integrates
    three core components: a comprehensive Python library featuring essential data
    processing and evaluation tools, along with leaderboards; an innovative \omnigen\ module
    that generates high-quality benchmarks from limited seed data; and \bench, a
    multi-layered benchmark suite covering {14} spectroscopic tasks and {over 10 spectrum types},
    {featuring spectra curated from over 1.2 million distinct chemical substances.}
    Thorough empirical studies on \bench\ with 23 cutting-edge multimodal
    LLMs reveal critical limitations of current approaches. We hope \lab\ will
    serve as a crucial foundation for future advancements in deep learning-driven
    spectroscopy. The anonymous code are available at \url{https://github.com/little1d/SpectrumLab}.
  \end{abstract}

  \section{Introduction}

  Spectroscopy, which investigates the interaction between electromagnetic radiation
  and matter, provides a powerful way to investigate the molecular structure and
  properties~\citep{elias_intensity-based_2004,prasad_review_2025}. By capturing
  characteristic patterns, such as peaks and shifts, in signals analogous to
  audio waveforms, spectroscopy offers a compact, information-rich
  representation of molecular systems~\citep{ralbovsky_towards_2020}. This low-dimensional
  encoding is indispensable in chemistry~\citep{silber_adsorbate-induced_2016,seo_infrared_2017},
  and life sciences~\citep{ralbovsky_towards_2020, zhang_molecular_2023, gasparin_label-free_2025}.
  It is not only central to molecular structure elucidation (\textit{i.e.}, Spectrum-to-Molecule
  structure) and property prediction, but also a key enabler for new material discovery
  and drug screening. In recent years, machine learning methods, especially deep
  learning, have demonstrated tremendous potential in spectroscopic data analysis,
  opening a new era of automation and intelligence in spectroscopy research~\citep{gastegger_machine_2017,gerrard_impression_2019, fine2020spectral, han2022concise, zou2023deep, devata2024deepspinn, lu_vib2mol_2025}.

  Despite recent advances, deep learning for spectroscopy still faces several
  fundamental challenges. Specifically, high-quality experimental spectral data
  remain scarce and expensive to acquire~\citep{Van_de_Sande2023-mq, Flanagan2025-tp},
  leading to public datasets that are limited in size and suffer from highly
  imbalanced distributions~\citep{bongiorno2022exploring,stenning2024neuromorphic,peng2025machine},
  which severely restricts model generalization. In addition, a substantial
  domain gap exists between experimental and computational spectra due to
  complex measurement conditions~\citep{agarwala2022experimental}, hindering the
  deployment of models trained on theoretical data. Furthermore, spectroscopy is
  inherently multimodal: it encompasses various spectral types (\textit{e.g.},
  infrared, Raman, nuclear magnetic resonance) represented as either 1D signals or
  2D images, often requiring integration with other molecular modalities such as
  molecular graphs, SMILES strings, and 3D conformations~\citep{litsa2021spec2mol,devata2024deepspinn}.
  The heterogeneous nature and semantics of these data modalities pose
  significant challenges for deep learning systems. Finally, the field lacks
  standardized benchmarks, with a fragmented landscape of tasks and datasets making
  it difficult to systematically evaluate and compare model performance.

  To address these challenges, we introduce \lab, a modular platform
  that streamlines the entire lifecycle of AI-driven spectroscopy from data
  preprocessing to model evaluation. Built on top of \lab, we construct
  \bench, a unified benchmark suite designed to evaluate machine learning
  models across diverse spectroscopic tasks and modalities. In contrast to
  existing approaches such as DiffSpectra~\citep{wang2025diffspectra} and MolSpectra~\citep{wang2025molspectra},
  which rely on contrastive learning and diffusion architectures, we are among
  the first to incorporate multi-modal large language models (MLLMs) into
  spectroscopic learning, using their alignment capabilities to bridge heterogeneous
  data modalities.

  \begin{figure}[!htb]
    \centering
    \includegraphics[width=0.8\linewidth]{
      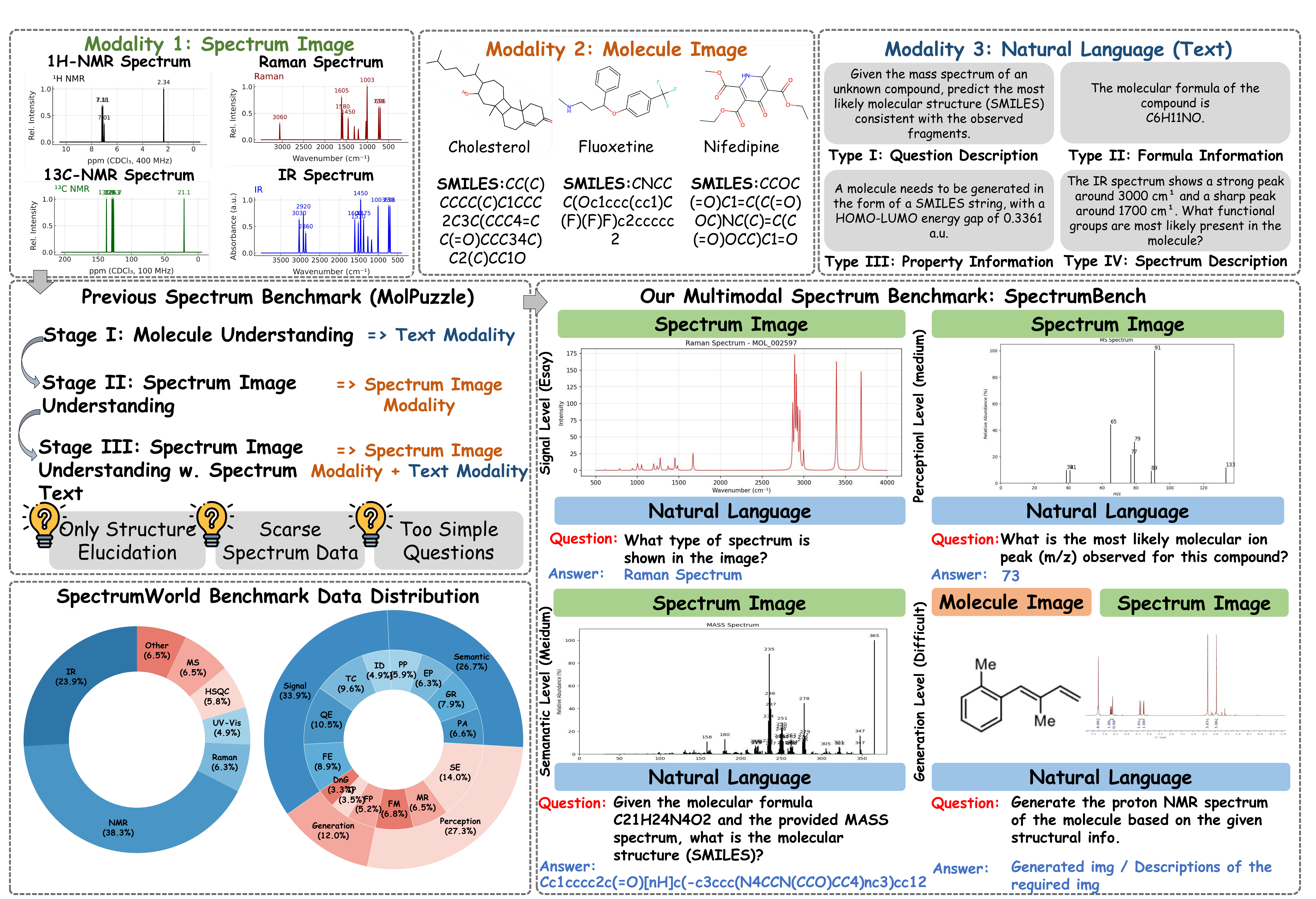
    }
    \caption{Overview of \bench. }
    \label{fig:SpectrumBench}
  \end{figure}

  \noindent
  Our main contributions are:
  \noindent
  (1) We introduce \textbf{\lab}, the first standardized framework tailored for spectroscopic
  machine learning with multimodal large language models, enabling reproducible
  pipelines from raw spectra to evaluation.
  \noindent
  (2) We design \textbf{\omnigen}, an automatic benchmark generator that constructs
  task-specific datasets from spectrum seeds, greatly accelerating prototyping
  and stress-testing of new models.
  \noindent
  (3) We release \textbf{\bench}, a large-scale benchmark suite covering diverse
  spectroscopic modalities and tasks, accompanied by unified evaluation
  protocols and public leaderboards to foster fair comparison and community
  progress.


  \section{Related Work}
  \begin{figure*}[!htb]
    \caption{Representative SpectraML methods categorized by \textbf{\textcolor[RGB]{16,70,128}{Spectral
    Type (left Y-axis)}} and \textbf{\textcolor[RGB]{109,1,31}{Model Type (right
    Y-axis)}}. Each dot indicates the use of a specific spectral modality or model
    architecture in a given method. Note that Raman is not included; thus,
    methods using it (\textit{e.g.}, DeepCID \citep{fan2019deep}) are not shown on
    the left Y-axis.}
    \footnotesize
    \centering
    \begin{overpic}
      [width=1\linewidth]{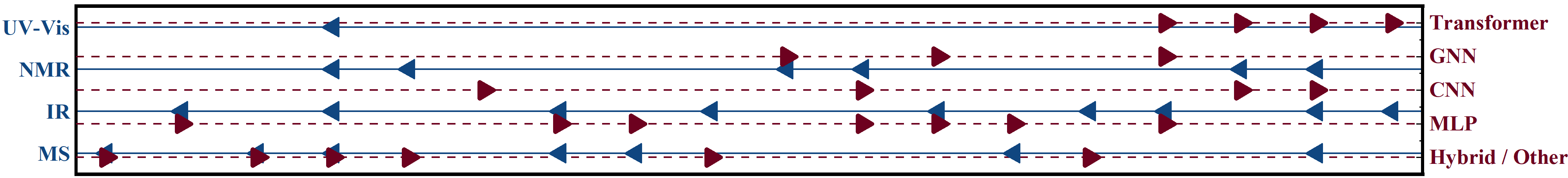}

      \put(0.5,0.4){ \rotatebox{-35}{ \fontsize{5}{5}\selectfont \begin{minipage}{1.8cm}\centering \textbf{CompLOD} \\ \cite{elias_intensity-based_2004}\end{minipage} } }

      \put(5.4,0.4){ \rotatebox{-35}{ \fontsize{5}{5}\selectfont \begin{minipage}{1.8cm}\centering \textbf{HDNNP} \\ \cite{gastegger2017machine}\end{minipage} } }
      \put(10.8,0.4){ \rotatebox{-35}{ \fontsize{5}{5}\selectfont \begin{minipage}{1.8cm}\centering \textbf{pDeep} \\ \cite{zhou2017pdeep}\end{minipage} } }

      \put(15.8,0.4){ \rotatebox{-35}{ \fontsize{5}{5}\selectfont \begin{minipage}{1.8cm}\centering \cite{de2018tutorial}\end{minipage} } }

      \put(20.8,0.4){ \rotatebox{-35}{ \fontsize{5}{5}\selectfont \begin{minipage}{1.8cm}\centering \textbf{ShiftML} \\ \cite{paruzzo2018chemical}\end{minipage} } }
      \put(25.3,0.4){ \rotatebox{-35}{ \fontsize{5}{5}\selectfont \begin{minipage}{1.8cm}\centering \textbf{DeepCID} \\ \cite{fan2019deep}\end{minipage} } }

      \put(30.1,0.4){ \rotatebox{-35}{ \fontsize{5}{5}\selectfont \begin{minipage}{1.8cm}\centering \cite{fine2020spectral}\end{minipage} } }
      \put(35.3,0.4){ \rotatebox{-35}{ \fontsize{5}{5}\selectfont \begin{minipage}{1.8cm}\centering \textbf{DeepEI} \\ \cite{ji2020predicting}\end{minipage} } }
      \put(40.3,0.4){ \rotatebox{-35}{ \fontsize{5}{5}\selectfont \begin{minipage}{1.8cm}\centering \cite{ren2021machine}\end{minipage} } }
      \put(44.8,0.4){ \rotatebox{-35}{ \fontsize{5}{5}\selectfont \begin{minipage}{1.8cm}\centering \textbf{ExpNN-ff} \\ \cite{Guan2021-jr}\end{minipage} } }

      \put(49,0.4){ \rotatebox{-35}{ \fontsize{5}{5}\selectfont \begin{minipage}{1.8cm}\centering \cite{huang2021framework}\end{minipage} } }
      \put(53.5,0.4){ \rotatebox{-35}{ \fontsize{5}{5}\selectfont \begin{minipage}{1.8cm}\centering \textbf{Chemprop-IR} \\ \cite{mcgill2021predicting}\end{minipage} } }

      \put(58,0.4){ \rotatebox{-35}{ \fontsize{5}{5}\selectfont \begin{minipage}{1.8cm}\centering \textbf{MS2DeepScore} \\ \cite{huber2021ms2deepscore}\end{minipage} } }
      \put(63.3,0.4){ \rotatebox{-35}{ \fontsize{5}{5}\selectfont \begin{minipage}{1.8cm}\centering \textbf{HD-NNP} \\ \cite{chen2024accelerating}\end{minipage} } }
      \put(68.3,0.4){ \rotatebox{-35}{ \fontsize{5}{5}\selectfont \begin{minipage}{1.8cm}\centering \textbf{Graphormer-IR} \\ \cite{stienstra2024graphormer}\end{minipage} } }
      \put(74,0.4){ \rotatebox{-35}{ \fontsize{5}{5}\selectfont \begin{minipage}{1.8cm}\centering \cite{hu2024accurate}\end{minipage} } }
      \put(78.3,0.4){ \rotatebox{-35}{ \fontsize{5}{5}\selectfont \begin{minipage}{1.8cm}\centering \cite{DBLP:conf/nips/AlbertsSZHL24}\end{minipage} } }
      \put(83,0.4){ \rotatebox{-35}{ \fontsize{5}{5}\selectfont \begin{minipage}{1.8cm}\centering \cite{alberts2024leveraging}\end{minipage} } }
    \end{overpic}
    \label{figure:related_work_SpectraML}
  \end{figure*}
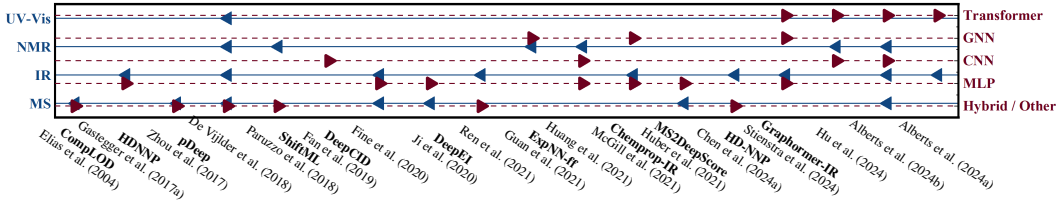

  \noindent
  \textbf{Machine Learning for Spectroscopy}. Spectroscopy is fundamental for molecular
  structure analysis and scientific discovery, enabling insights into chemical
  properties and interactions~\citep{DBLP:journals/corr/abs-2502-09897}. Its
  applications span diverse scientific domains, including chemistry, material
  science, and drug development~\citep{Shao2025-uq, DBLP:journals/corr/abs-2502-15867}.
  Machine learning techniques have been extensively applied in spectroscopy for tasks
  such as molecular structure elucidation (spectrum-to-molecule)~\citep{kuhn2008building, de2018tutorial, paruzzo2018chemical, fan2019deep, nguyen2019recent, ji2020predicting, fine2020spectral, huang2021framework, hu2024accurate, lu_vib2mol_2025, liu2025current}
  and spectral simulation (molecule-to-spectral)~\citep{gastegger2017machine, zhou2017pdeep, liu2017deep, mcgill2021predicting, Guan2021-jr, ren2021machine, DBLP:journals/natmi/YoungRW24}.
  As illustrated in Figure~\ref{figure:related_work_SpectraML}, recent efforts have
  explored a variety of spectral modalities, such as IR~\citep{hu2024accurate},
  NMR~\citep{hu2024accurate}, UV-Vis~\citep{de2018tutorial}, MS~\citep{huber2021ms2deepscore},
  and Raman~\citep{fan2019deep}, and have adopted heterogeneous deep learning
  model architectures, ranging from MLPs~\citep{stienstra2024graphormer} and
  CNNs~\citep{DBLP:conf/nips/AlbertsSZHL24} to GNNs~\citep{mcgill2021predicting}
  and Transformers~\citep{alberts2024leveraging}. Despite these rapid progresses,
  existing methods still face several limitations: {(1)} most studies are
  constrained to a single modality (\textit{e.g.}, IR or MS), lacking generalization
  across spectral types~\citep{Beck2024-lu}; {(2)} the field lacks unified benchmarks
  and evaluation protocols, making objective comparisons difficult; {(3)} dataset
  sizes remain limited and imbalanced, further impeding reproducibility and
  robustness; {(4)} previous benchmarks does not support multi-modal large language
  models. These limitations highlight the need for standardized, cross-modal frameworks
  to advance machine learning for spectroscopy, especially spectroscopy
  foundation models.

  \noindent
  \textbf{Spectroscopy Foundation Models}. While foundation models have shown
  promising progress in scientific discovery~\citep{DBLP:journals/corr/abs-2505-16326, DBLP:journals/corr/abs-2502-07527},
  spectroscopy foundation models are still underexplored. This is largely due to
  the inherent multimodal nature of spectroscopic data, which combines spectral
  signals with diverse molecular representations. Although recent efforts such as
  SpectraFM~\citep{koblischke2024spectrafm} and LSM1-MS2~\citep{asher2024lsm1} have
  introduced pre-trained foundation models on Stellar and MS spectra for chemical
  property prediction, these models remain fundamentally single-modal, focusing solely
  on spectral information. Despite these challenges, the integration of spectroscopy
  into the foundation model paradigm holds significant promise for advancing automated
  analysis and multi-modal scientific discovery in the future.
  \begin{table*}
    [htbp]
    \caption{Comparison of Benchmark Studies. \textbf{Notes:} ``Other'' in the
    Spectral Modality column includes modalities not explicitly listed, such as HSQC
    (Heteronuclear single quantum coherence spectroscopy) and UV-Vis (Ultraviolet-visible
    spectroscopy). The NMR column refers to both $^{1}$H-NMR and $^{13}$C-NMR. We
    unify tasks' terminology for clarity.}
    \vspace{0.3cm}
    \label{tab:benchmark_comparison}
    \centering
    \begin{adjustbox}
      {max width=\textwidth}
      \begin{tabular}{llcccccccccccc}
        \toprule                                                                               
        \textbf{Benchmark}                                                                    & \textbf{Reference}                                                                  & \multicolumn{5}{c}{\textbf{Spectral Modality}}                                              & \multicolumn{7}{c}{\textbf{Task}} \\
        \cmidrule(lr){3-7} \cmidrule(lr){8-14}                                                 
                                                                                              &                                                                                     & \textbf{Raman}                                                                              & \textbf{IR}                      & \textbf{NMR} & \textbf{MS} & \textbf{Other} &             
        \multirow{2}{*}{\parbox[c]{2cm}{\centering \textbf{Molecular}\\\textbf{Elucidation}}} & \multirow{2}{*}{\parbox[c]{2cm}{\centering \textbf{Spectrum}\\\textbf{Simulation}}} & \multirow{2}{*}{\parbox[c]{2cm}{\centering \textbf{\textit{De novo}}\\\textbf{Generation}}} &                                   
        \multicolumn{4}{c}{\textbf{Understanding}}                                             \\
        \cmidrule(lr){11-14}                                                                   
                                                                                              &                                                                                     &                                                                                             &                                  &              &             &                &            &            &            & \textbf{GR} & \textbf{PA} & \textbf{FM} & \textbf{MR} \\
        \midrule NovoBench                                                                    & \citep{DBLP:journals/corr/abs-2406-11906}                                           &                                                                                             &                                  &              & \checkmark  &                &            &            & \checkmark &             &             &             &             \\
        MolPuzzle                                                                             & \citep{NEURIPS2024_f2b9e8e7}                                                        &                                                                                             & \checkmark                       & \checkmark   & \checkmark  &                & \checkmark &            &            & \checkmark  &             &             &             \\
        Multimodal Spec                                                                       & \citep{DBLP:conf/nips/AlbertsSZHL24}                                                &                                                                                             & \checkmark                       & \checkmark   & \checkmark  & \checkmark     & \checkmark & \checkmark &            & \checkmark  &             &             &             \\
        MassSpecGym                                                                           & \citep{DBLP:conf/nips/BushuievBJYKSHW24}                                            &                                                                                             &                                  &              & \checkmark  &                & \checkmark & \checkmark &            &             &             &             &             \\
        NMRNet                                                                                & \citep{Xu2025-qr}                                                                   &                                                                                             &                                  & \checkmark   &             &                &            &            &            &             & \checkmark  &             &             \\
        ViBench                                                                               & \citep{lu_vib2mol_2025}                                                             & \checkmark                                                                                  & \checkmark                       &              &             &                & \checkmark &            &            &             &             &             &             \\
        {\bench}                                                                              & Ours                                                                                & \checkmark                                                                                  & \checkmark                       & \checkmark   & \checkmark  & \checkmark     & \checkmark & \checkmark & \checkmark & \checkmark  & \checkmark  & \checkmark  & \checkmark  \\
        \bottomrule[0.9pt]
      \end{tabular}
    \end{adjustbox}
    \begin{minipage}{\textwidth}
      \footnotesize \textbf{Abbreviations:} GR = Functional Group Recognition, PA
      = Peak Assignment, FM = Fusing Spectroscopic Modalities, MR = Multimodal Molecular
      Reasoning.
    \end{minipage}
  \end{table*}

  \textbf{Benchmark and Toolkits for Spectroscopy}. Several benchmarks and
  toolkits have been developed to support spectroscopic machine learning
  research~\citep{heid2023chemprop, NEURIPS2024_bd281779, NEURIPS2024_c6c31413, NEURIPS2024_f2b9e8e7, devata2024deepspinn, ruan2024automatic, DBLP:journals/corr/abs-2502-09897}.
  However, many of these efforts remain limited in scope (either spectrum
  modalities or tasks), lacking extensibility and comprehensive evaluation
  across diverse spectroscopic tasks and modalities. For example, MassSpecGym~\citep{NEURIPS2024_c6c31413}
  focuses solely on MS data and does not incorporate language descriptions,
  hindering support for multi-modal inputs. Although MolPuzzle~\citep{NEURIPS2024_f2b9e8e7}
  enables multi-modal inputs, it omits Raman spectra and lacks support for pure spectral
  understanding tasks. Furthermore, several toolkits~\citep{NEURIPS2024_c6c31413, NEURIPS2024_bd281779}
  do not provide interfaces for multi-modal large language models (MLLMs), and even
  MolPuzzle lacks benchmarking for more recent MLLMs. In contrast, our \lab\ is a
  unified, extensible, and reproducible platform that addresses these limitations
  by supporting a wide range of spectroscopic tasks, modalities, and integration
  with MLLMs. Table~\ref{tab:benchmark_comparison} systematically compares
  representative studies in terms of their spectral modality and task coverage. \lab\ not
  only fills critical gaps in data, evaluation, and tooling, but also
  establishes a new standard for spectroscopic AI and enables future advances in
  multi-modal, large-model-driven scientific discovery.


  \section{\bench}
  \noindent
  \textbf{Overview.} \bench\ is a unified benchmark suite for deep learning in spectroscopy,
  covering four hierarchical levels and 14 sub-tasks that span from spectroscopy
  understanding to generation. All questions and tasks are initially defined by domain
  experts, and subsequently refined and validated through expert review and
  rigorous quality assurance processes. Compared
  to existing benchmarks, \bench\ offers broad modality and task coverage within
  a standardized, extensible framework for fair and reproducible model evaluation.
  
\begin{wraptable}[16]{r}{0.5\textwidth} 
  \caption{Tasks' categories and statistics.}
  \label{tab:task_distribution}
  \centering
  \begin{adjustbox}{max width=0.5\textwidth}
    \footnotesize
    \begin{tabular}{@{}llc@{}}
      \toprule
      \textbf{Category} & \textbf{Task} & \textbf{\# questions} \\
      \midrule
      \multirow{4}{*}{Signal} & Spectrum Type Classification (TC) & 55 \\
      & Spectrum Quality Assessment (QE) & 60 \\
      & Basic Feature Extraction (FE) & 51 \\
      & Impurity Peak Detection (ID) & 28 \\
      \midrule
      \multirow{4}{*}{Perception} & Functional Group Recognition & 45 \\
      & Elemental Compositional Prediction (EP) & 36 \\
      & Peak Assignment (PA) & 38 \\
      & Basic Property Prediction (PP) & 34 \\
      \midrule
      \multirow{3}{*}{Semantic} & Molecular Structure Elucidation (SE) & 80 \\
      & Fusing Spectroscopic Modalities (FM) & 39 \\
      & Multimodal Molecular Reasoning (MR) & 37 \\
      \midrule
      \multirow{3}{*}{Generation} & Forward Problems (FP) & 30 \\
      & Inverse Problems (IP) & 20 \\
      & \textit{De Novo} Generation (DnG) & 19 \\
      \bottomrule
    \end{tabular}
  \end{adjustbox}
\end{wraptable}

  \noindent
  \textbf{Spectroscopic Type.} Unlike previous benchmarks that are limited to a
  single spectroscopic modality or narrowly defined data types \citep{DBLP:conf/nips/BushuievBJYKSHW24},
  \bench\ integrates a diverse array of spectroscopic data sources. Our \bench\ benchmark
  currently includes more than 10 distinct types of spectroscopic data, such as
  infrared (IR), nuclear magnetic resonance (NMR), and mass spectrometry (MS). As
  illustrated in Figure~\ref{fig:SpectrumBench},
  this comprehensive data foundation accurately reflects the diverse and complex
  multi-modal spectroscopic scenarios encountered in real-world applications.

  \noindent
  \textbf{Task.} In contrast to previous benchmarks that primarily focus on
  molecule elucidation or spectrum simulation, \bench\ encompasses a much broader
  spectrum of task types. \bench\ is organized according to a multi-level
  hierarchical taxonomy that systematically covers tasks ranging from low-level
  signal analysis to high-level semantic reasoning and generative challenges. This
  taxonomy, developed through expert consultation and iterative refinement, comprises
  four principal layers: \textbf{signal, perception, semantic, and generation}.
  Each layer is further divided into several subcategories, capturing a diverse set
  of scientific and application-driven tasks. Detailed definitions and
  representative examples for each task layer are provided in the Appendix
  \ref{sec:BenchDetailedInfo}.

  \subsection{Data Curation Pipeline}

  \begin{figure*}[htbp]
    \centering
    \includegraphics[width=0.8\textwidth]{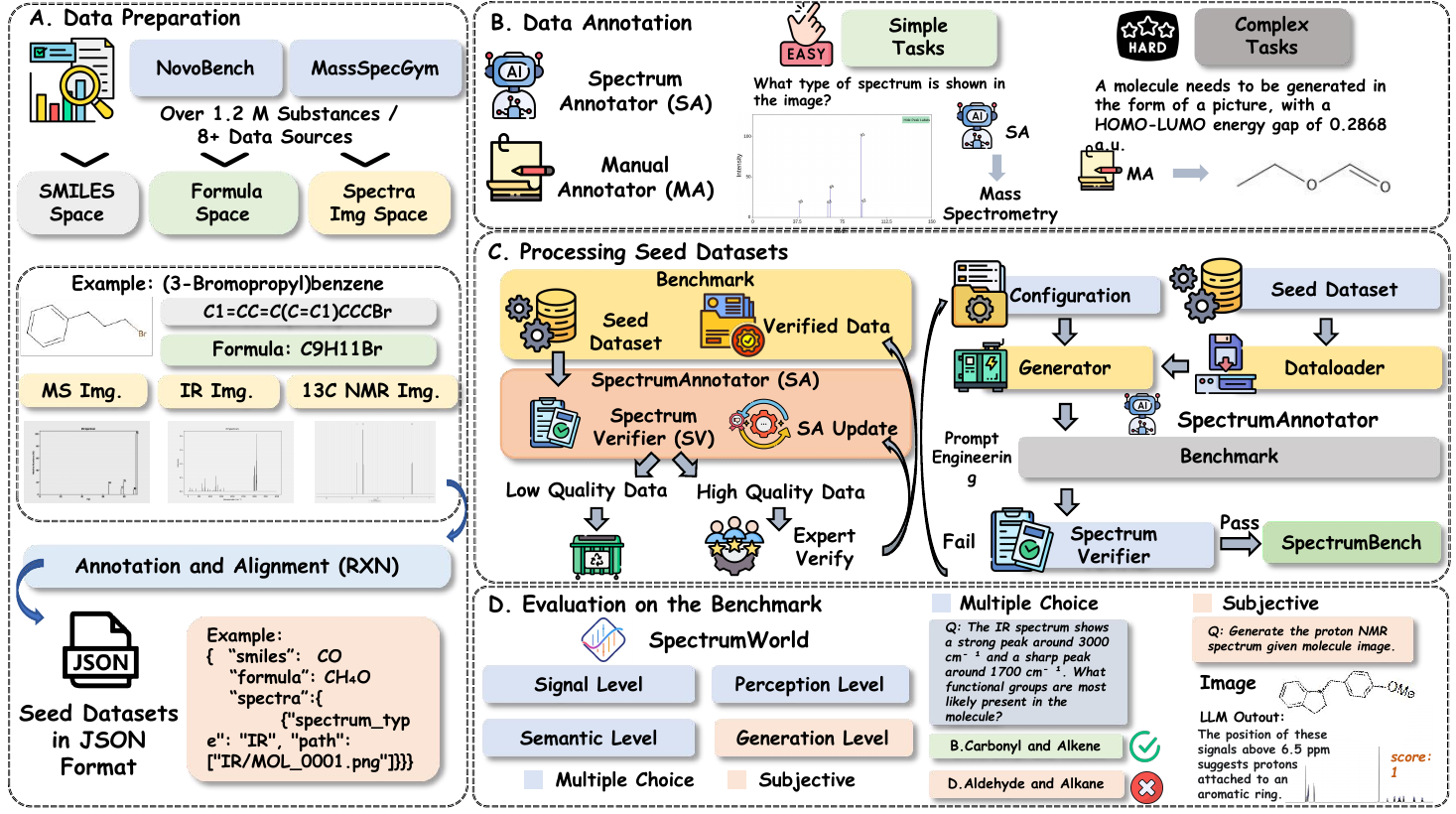}
    \caption{Overview of the data curation pipeline used in \bench. }
    \label{fig:data_curation}
  \end{figure*}

  \noindent
  \textbf{Task Construction.} Spectroscopic machine learning encompasses a wide spectrum
  of tasks, driven by the intrinsic complexity of molecular structures and the
  multifaceted nature of spectroscopic data. These tasks often involve diverse input
  modalities (e.g., molecular graphs, SMILES, textual prompts) and equally varied
  outputs (e.g., spectra, chemical attributes, structured predictions), which
  reflect the real-world demands of chemical analysis, property reasoning, and
  molecular generation. To illustrate this diversity, we organize existing
  spectroscopic tasks into four broad input-output categories:
  (1) \textit{Molecule-to-Spectrum (Spectrum Simulation)} aims at generating a spectrum
  based on molecular structure.
  (2) \textit{Spectrum-to-Molecule (Molecule Elucidation)} refers to the tasks
  that infer molecular structures from spectra.
  (3) \textit{Text-to-Any\footnote{``Any'' encompasses various data modalities, such
  as molecular representations, spectral data, or peptide sequences.} (\textit{De
  novo} Generation)} refers to the task of generating novel, diverse, and reasonable
  molecular structures (SMILES string, 2D molecular graph) and/or predicting
  multimodal information (spectra, properties) according to specific goals (\textit{e.g.},
  molecules of a specific nature, ligands of a specific target).
  Moreover, in previous studies, many tasks involving inferring molecular
  structures from spectra were also categorized under ``\textit{de novo} generation''~\citep{NEURIPS2024_c6c31413,lu_vib2mol_2025}.
  While this has some rationality, for the sake of consistency in our task framework,
  we clarify that our defined \textit{de novo} generation task has distinct characteristics:
  its input consists solely of textual descriptions, which may include specifications
  of molecular properties (\textit{e.g.}, desired chemical natures, target-binding
  affinities), without involving spectral data as input. Meanwhile, the output scope
  is broader, encompassing not only molecular structures but also spectral and
  textual descriptions of molecules.
  (4) \textit{Any-to-Text (Understanding)}.
  Tasks in signal, perception, semantic layers fall under the ``Understanding''
  category. Its task type is presented in the form of multiple-choice questions,
  which may include tasks such as inferring the molecular structure from
  spectrum (\textit{e.g.}, functional group recognition, peak attribution tasks).
  This partially overlaps with the molecular elucidation tasks described above. For
  a compromise design, we use the output form to distinguish between them. The question
  format of ``Understanding'' tasks will only be multiple-choice questions,
  which means the output is text.

  \noindent
  \textbf{Taxonomy Definition}.
  These input-output patterns offer a high-level overview of the task landscape.
  However, previous works often cover only a subset of them, limiting both their generalizability and their ability to benchmark diverse ML capabilities. We show these patterns in Table~\ref{tab:benchmark_comparison}, which highlights substantial heterogeneity
between existing methods. To address this limitation and support more
structured and extensible benchmarking, we propose a four-level hierarchical taxonomy tailored to spectroscopic machine learning: \textit{Signal}, \textit{Perception},
  \textit{Semantic}, and \textit{Generation}—is designed to reflect the logic of real-world scientific workflows in spectroscopy. As depicted in Figure~\ref{fig:data_curation},
  this layered structure systematically provides a robust framework for our 14 meticulously
  designed tasks detailed in Table~\ref{tab:task_distribution}.

  \noindent
  (1) \textit{Signal level:} This foundational layer focuses on the direct
  analysis and processing of raw spectral data, such as spectrum type classification
  and peak detection. Tasks at this level are designed to extract and refine
  primary features from experimental measurements, mirroring the initial steps taken
  by chemists to prepare and interpret spectra in the laboratory. This level
  primarily encompasses Any-to-Text(Understanding) tasks that operate directly on
  raw signal data.

  \noindent
  (2) \textit{Perception level:} Building upon the processed signals, the perception
  layer addresses pattern recognition and intermediate interpretation tasks,
  such as functional group identification, peak assignment, and basic molecule
  properties prediction. This stage reflects the chemist's effort to translate spectral
  features into meaningful chemical information, bridging the gap between raw
  data and higher-level understanding. Many Any-to-Text(Understanding) tasks
  that involve interpreting specific patterns within spectra fall into this
  category.

  \noindent
  (3) \textit{Semantic level:} At this layer, the focus shifts to comprehensive
  molecular reasoning and property inference, including molecule elucidation and
  cross-modal correlation (\textit{e.g.}, linking spectra to molecular graphs or
  textual descriptions). The semantic layer encapsulates the core scientific
  reasoning that underpins hypothesis generation and validation in spectroscopic
  research, primarily addressing advanced Any-to-Text(Understanding) tasks that
  require intricate chemical knowledge and contextualization.

  \noindent
  (4) \textit{Generation level:} The final layer encompasses creative and generative
  tasks, where new entities are produced. The level explicitly consolidates all tasks
  involving the synthesis of new data or structures, including Molecule-to-Spectrum
  (\textit{e.g.}, direct spectrum generation from molecular inputs), Spectrum-to-Molecule
  (\textit{e.g.}, generates a molecular structure from spectra input). These
  tasks emulate advanced scientific workflows where new hypotheses, molecules,
  or spectral data are generated to drive discovery and innovation.

  \begin{figure}
      \centering
      \includegraphics[width=0.8\linewidth]{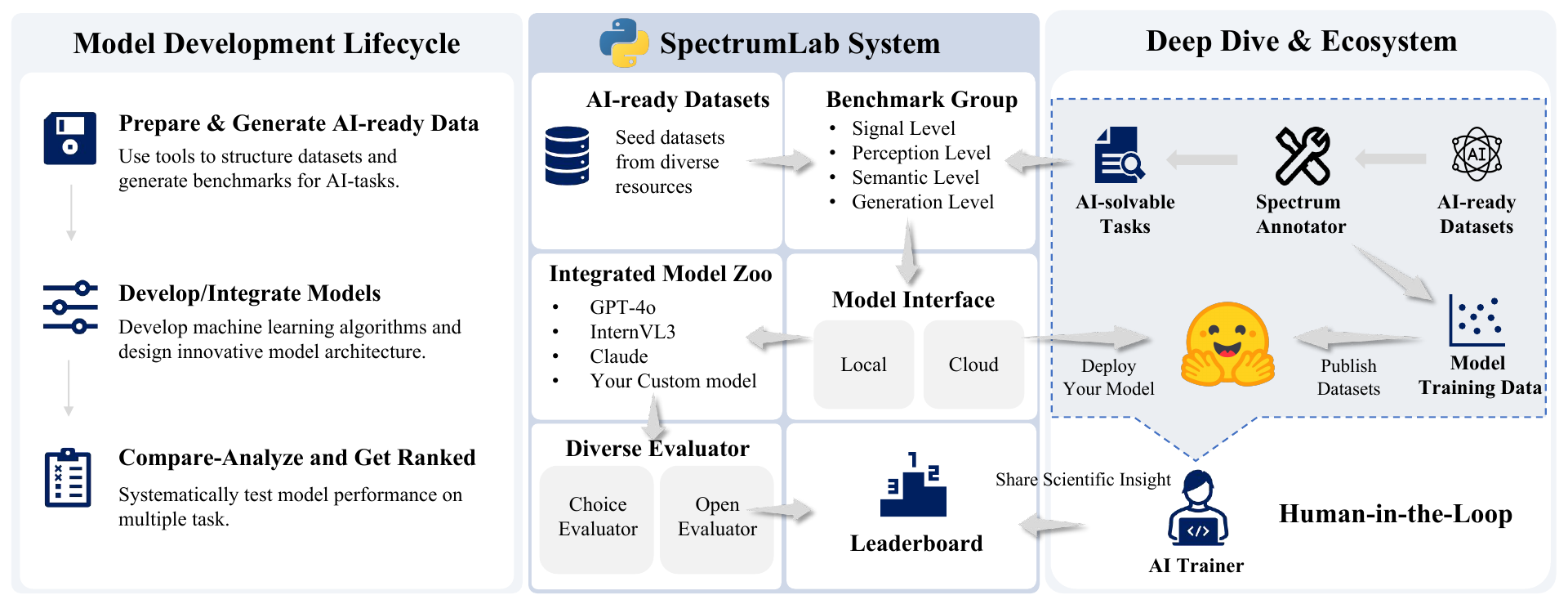}
      \caption{SpectrumLab is the main component of \name, which could be extended conveniently. More multi-modal spectrum data could be included, as well as MLLMs. 
      }
      \label{fig:spectrumlan}
  \end{figure}

  \noindent
\textbf{Seed Data Preparation}. The seed datasets for this work were curated from three primary sources: proprietary experimental data, public repositories, and literature mining. As illustrated in Figure~\ref{fig:data_curation}, the construction of seed datasets begins with the aggregation of raw data from multiple sources. All collected datasets undergo a unified processing pipeline that systematically maps each entry into three core chemical spaces: SMILES string, molecular formula, and spectra. Rigorous cleaning, normalization, and deduplication ensure data consistency and reliability. Following this, human annotation and alignment are performed to guarantee scientific accuracy and completeness. The resulting seed datasets are organized at the level of individual chemical substances, with each record containing the compound's SMILES, molecular formula, and a structured set of associated spectra, all stored in a standardized JSON format to facilitate downstream annotation. Detailed descriptions of the seed datasets and the standardization process are provided in Appendix~\ref{sec:DataStructure}. We ensure that the curated seed data are not contaminated.

  \noindent
  \textbf{Data Annotation}. We use two annotation methods: automated and manual annotation.
  \textit{(1) Automated annotation (\omnigen)}. For tasks characterized by well-defined
  rules like spectrum recognition
  , we design \omnigen,
  a core contribution of this work. \omnigen is a novel, self-developed annotation
  framework that harnesses the zero-shot and multi-modal reasoning capabilities of
  state-of-the-art MLLMs. Given curated seed datasets and a set of pre-defined
  benchmark prompts, \omnigen\  automatically designs and generates high-quality,
  multi-modal benchmark data.
  Further technical details and implementation specifics of \omnigen\ are
  provided in Appendix~\ref{sec:Annotator}. \textit{(2) Manual Annotation}. For
  more complex or open-ended tasks, particularly those involving multi-step reasoning
  or sophisticated scientific interpretation, manual annotation by domain
  experts is indispensable. Human annotators ensure the scientific validity and depth
  of the benchmark, especially in cases where automated methods cannot handle. 

  \noindent
  \textbf{Data Quality Assurance}. To ensure the integrity and reliability of
  \bench, we implement a comprehensive quality assurance pipeline, as
  illustrated in Figure~\ref{fig:data_curation}. The process begins with \textit{Candidate
  Data} undergoing automated screening by the \textit{SpectrumVerifier} (\textbf{SV}).
  This stage efficiently detects and filters out clear errors such as missing options
  or image-text discrepancies, categorizing them as \textit{Low Quality Data} for
  removal. Remaining \textit{High Quality Data} proceeds to expert manual evaluation.
  If issues are identified, a feedback loop through internal annotator update initiates
  targeted reannotation via \textit{SpectrumAnnotator} (\textbf{SA}). This multi-stage
  quality control ensures only high-quality, scientifically robust data are included
  in our final benchmark.

  \section{\lab}

  \subsection{System Overview}

  \noindent
  \textbf{AI-ready Datasets and AI-solvable Tasks}. \lab\ is tightly integrated with
  \omnigen, which is responsible for generating high-quality benchmarks from
  seed datasets collected from diverse sources. In this workflow, \omnigen\
  curates a wide range of scientifically rigorous benchmarks from these seeds. \lab\ then
  offers a flexible abstraction for users to define and encapsulate specific AI-solvable
  tasks based on these curated benchmarks. A core abstraction unique to \lab\ is
  the \textit{Benchmark Group}. Users can combine multiple benchmark instances
  or select specific subsets to form a \textit{Benchmark Group}, creating
  tailored task definitions within a unified framework. By encapsulating benchmarks
  as tasks, \lab\ streamlines the process of task definition and evaluation.

  \noindent
  \textbf{Toolkits and Ecosystem}. \lab\ offers a flexible ecosystem of Python libraries
  and tools designed to streamline the entire workflow for spectroscopy, from
  data preprocessing to model evaluation. Its modular design allows seamless integration
  of custom models and tasks. Distributed via the Python Package Index(PyPI) for
  easy installation, \lab\ provides a comprehensive environment for state-of-the-art
  machine learning research in spectroscopy.

  \noindent
  \textbf{Leaderboards}. To ensure transparency and reproducibility, \lab\ incorporates
  a comprehensive public leaderboard system that systematically tracks and compares
  the performance of a wide range of models across all tasks. The leaderboard
  provides fine-grained reporting, recording each model's results on both high-level
  and detailed tasks. The platform currently supports benchmarking for over 20
  MLLMs, including prominent open-source models such as InternVL3~\citep{DBLP:journals/corr/abs-2504-10479}
  and proprietary models like GPT-4o~\citep{DBLP:journals/corr/abs-2410-21276},
  across 14 specific tasks.

  \subsection{Modular Design}

  \begin{table*}
    [!htb]
    \centering
    \begin{adjustbox}
      {max width=\textwidth}
      \large
      \begin{tabular}{l||*{4}{C}|*{4}{C}|*{3}{C}|*{3}{C}|CC}
        \toprule \textbf{Model}                                                                         & \multicolumn{4}{c|}{\textbf{Signal}} & \multicolumn{4}{c|}{\textbf{Perception}} & \multicolumn{3}{c|}{\textbf{Semantic}} & \multicolumn{3}{c}{\textbf{Generation}} & \multicolumn{2}{|c}{\textbf{Avg.}} \\
        \cmidrule(r){2-5} \cmidrule(r){6-9} \cmidrule(r){10-12} \cmidrule(r){13-15} \cmidrule(r){16-17} & \makecell{TC}                        & \makecell{QE}                            & \makecell{FE}                          & \makecell{ID}                           & \makecell{GR}                     & \makecell{EP}     & \makecell{PA}     & \makecell{PP}     & \makecell{SE}     & \makecell{FM}     & \makecell{MR}     & \makecell{FP}     & \makecell{IP}     & \makecell{DnG}    & Perf              \\
        \midrule

\rowcolor{gray!20} \multicolumn{17}{c}{\textit{Closed-source MLLMs}}                   \\
        \midrule Claude-3.5-Sonnet                                                                      & 96.36                                & 28.33                                    & 76.47                                  & 71.43                                   & 60.00                             & 77.78             & 76.32             & 85.29             & 82.50             & 69.23             & 94.59             & 20.00             & 0                 & 0                 & 59.88             \\
        Claude-3.7-Sonnet                                                                               & 96.36                                & \textbf{38.33}                           & 86.27                                  & 82.14                                   & 71.43                             & \textbf{88.89}    & 71.05             & \underline{88.24} & 82.28             & 74.36             & 89.19             & 20.00             & 0                 & 5.26              & 63.84             \\
        Claude-4-Sonnet                                                                                 & 96.36                                & 35.00                                    & \underline{88.24}                      & \textbf{92.86}                          & 62.22                             & 63.89             & 60.53             & 76.47             & 16.25             & 43.59             & 64.86             & 3.33              & 0                 & \underline{21.05} & 51.76             \\
        Claude-3.5-Haiku                                                                                & 94.55                                & 31.67                                    & 50.98                                  & \textbf{92.86}                          & 66.67                             & 75.00             & 76.32             & 76.47             & 67.50             & 64.10             & 81.08             & 10.00             & 0                 & 0                 & 56.23             \\
        Claude-4-Opus                                                                                   & 96.36                                & 33.33                                    & 86.27                                  & \textbf{92.86}                          & \underline{73.33}                 & 83.33             & 71.05             & 85.29             & 32.50             & \underline{76.92} & 86.49             & 16.67             & 0                 & 5.26              & 59.98             \\
        GPT-4o                                                                                          & 96.36                                & 33.33                                    & 68.63                                  & \textbf{92.86}                          & 59.88                             & 77.78             & 63.16             & 79.41             & 78.75             & 58.97             & 89.19             & 10.00             & 0                 & 0                 & 57.74             \\
        GPT-4.1                                                                                         & 94.55                                & 28.33                                    & 86.27                                  & 85.71                                   & 53.33                             & 77.78             & 63.16             & 79.41             & 82.50             & 66.67             & 91.89             & 33.33             & \underline{10.53} & 0                 & 60.96             \\
        GPT-4-Vision                                                                                    & 94.55                                & 33.33                                    & 72.55                                  & \textbf{92.86}                          & \underline{73.33}                 & 72.22             & 71.05             & 82.35             & 73.75             & 53.85             & \underline{97.30} & 23.33             & 5.00              & 0                 & 60.39             \\
        Gemini-2.5-pro                                                                                  & 96.36                                & 35.00                                    & \textbf{90.20}                         & 67.86                                   & \textbf{75.56}                    & \underline{86.11} & 65.79             & 79.41             & 68.75             & \textbf{84.62}    & \underline{97.30} & \underline{50.00} & 5.00              & \textbf{47.37}    & \textbf{67.81}    \\
        Grok-2-Vision                                                                                   & 94.55                                & 31.67                                    & 74.51                                  & \underline{89.29}                       & 64.44                             & 80.56             & 73.68             & 82.35             & 37.50             & 66.67             & 81.08             & 23.33             & 0                 & 0                 & 57.12             \\
        Qwen-VL-Max                                                                                     & 94.55                                & \underline{36.67}                        & \textbf{90.20}                         & \textbf{92.86}                          & 60.00                             & 80.56             & \underline{78.95} & \underline{88.24} & 32.50             & 71.79             & 91.89             & 43.33             & 0                 & 5.26              & 61.91             \\
        Doubao-1.5-Vision-Pro                                                                           & 98.18                                & 33.33                                    & 78.43                                  & \textbf{92.86}                          & 66.67                             & 83.33             & 68.42             & \underline{88.24} & 67.50             & 56.41             & 89.19             & 6.67              & 0                 & 0                 & 59.23             \\
        Doubao-1.5-Vision-Pro-Thinking                                                                  & 96.36                                & 35.00                                    & 78.43                                  & 67.86                                   & 53.33                             & 80.56             & 73.68             & \textbf{91.18}    & 68.75             & 66.67             & 91.89             & \textbf{66.67}    & 5.00              & 5.26              & 62.90             \\
        \rowcolor{gray!20} \multicolumn{17}{c}{\textit{Open-source MLLMs}}                               \\
        \midrule Qwen2.5-VL-32B-Instruct                                                                & 92.73                                & 26.67                                    & 37.25                                  & 71.43                                   & 57.78                             & 44.44             & 31.58             & 61.76             & 0.00              & 5.13              & 45.95             & 20.00             & 0                 & 0                 & 35.34             \\
        Qwen2.5-VL-72B-Instruct                                                                         & 94.55                                & \textbf{38.33}                           & 86.27                                  & \textbf{92.86}                          & 42.22                             & 80.56             & \underline{78.95} & \underline{88.24} & 66.25             & \underline{76.92} & 91.89             & 30.00             & 0                 & 10.53             & 62.68             \\
        InternVL3-78B                                                                                   & 96.36                                & \textbf{38.33}                           & 70.59                                  & 71.43                                   & 48.49                             & 75.00             & \textbf{81.58}    & \underline{88.24} & 62.50             & 69.23             & 83.78             & 23.33             & 0                 & 5.26              & 58.15             \\
        InternVL3.5-241B                                                                                & \underline{98.18}                    & 33.33                                    & \textbf{90.20}                         & \textbf{92.86}                          & 66.67                             & 77.78             & 71.05             & 85.29             & 86.25             & 61.54             & \textbf{100.00}   & 33.33             & 10.00             & 10.53             & \underline{65.50} \\
        Llama-3.2-11B-Vision-Instruct                                                                   & 34.55                                & 11.67                                    & 13.73                                  & 25.00                                   & 20.00                             & 41.67             & 15.79             & 29.41             & 7.50              & 5.13              & 21.62             & 0                 & 0                 & 0                 & 16.15             \\
        Llama-3.2-90B-Vision-Instruct                                                                   & 38.18                                & 10.00                                    & 35.29                                  & 25.00                                   & 17.78                             & 27.78             & 28.95             & 20.59             & 21.25             & 5.13              & 43.24             & 0                 & 0                 & 0                 & 19.51             \\
        DeepSeek-VL2                                                                                    & 52.73                                & 23.33                                    & 29.41                                  & 28.57                                   & 8.89                              & 27.78             & 28.95             & 50.00             & 15.00             & 15.38             & 32.43             & 10.00             & 5.00              & 5.26              & 23.77             \\
        GLM-4.5V                                                                                        & \textbf{100.00}                      & 28.33                                    & 70.59                                  & 92.86                                   & \underline{73.33}                 & 83.33             & 71.05             & \textbf{91.18}    & 63.75             & 69.23             & 83.78             & 0                 & 0                 & 10.53             & 59.85             \\
        InternS1-nothink                                                                                & \underline{98.18}                    & \underline{36.67}                        & 72.55                                  & \underline{89.29}                       & 51.11                             & 72.22             & 73.68             & 79.41             & \underline{86.25} & 66.67             & 94.59             & 13.33             & 0                 & 0                 & 59.57             \\
        InternS1-think                                                                                  & \underline{98.18}                    & 25.00                                    & 80.39                                  & \underline{89.29}                       & 64.44                             & \textbf{88.89}    & 73.68             & \textbf{91.18}    & \textbf{90.00}    & 56.41             & 91.89             & 10.00             & \textbf{40.00}    & 15.79             & 65.37             \\
        \midrule \rowcolor{gray!10} \textbf{Overall Avg.}                                               & 89.09                                & 30.65                                    & 70.16                                  & 77.95                                   & 56.13                             & 71.62             & 63.84             & 76.85             & 56.08             & 55.85             & 79.79             & 20.29             & 3.50              & 6.41              & 54.15             \\
        \bottomrule[0.8pt]
      \end{tabular}
    \end{adjustbox}
    \caption{Accuracies (\%, $\uparrow$) of all models on different levels. Task
    abbreviations (e.g., TC, QE, FE, etc.) are defined in Table~\ref{tab:task_distribution}.
    \textbf{best: bold}, \underline{second best: underlined}. The second last column
    calculates the arithmetic mean and the last column calculates the true weighted
    mean of each row.}
    \label{tab:vlm-tasks}
  \end{table*}

  \lab\ adopts a modular architecture to maximize flexibility and extensibility.
  The core components include:

  \noindent
  (1) \textbf{Benchmark Group:} \lab\ organizes \bench\ into hierarchical
  groups corresponding to different levels of spectroscopic reasoning. This
  structure enables
  systematic evaluation across various tasks and spectroscopic modalities, while also supporting rapid assessment of specialized models on domain-specific spectra.
  spectra and tasks.

  \noindent
  (2) \textbf{Model Integration:} \lab\ offers a unified, extensible framework
  for integrating external models. Using standardized APIs and modular
  adapters, it connects seamlessly to a wide range of model types, from cloud-based services to locally deployed solutions, enabling consistent
  benchmarking within a single evaluation environment.

  \noindent
  (3) \textbf{Evaluator:} Serving as the abstract core of the benchmark
  evaluation engine, the Evaluator module in \lab\ is designed for flexible and extensible
  assessment of model performance across diverse spectroscopic tasks. It enables
  the customization of evaluation metrics and protocols according to the
  specific requirements of each task, and can be seamlessly integrated with both
  the \textit{Benchmark Group} and external model modules. This modular abstraction
  allows researchers to define and implement tailored evaluation strategies,
  ensuring rigorous and task-appropriate benchmarking. Currently, \lab\ supports
  the following two types of evaluators: (i) \textit{Choice Evaluator:}
  Specially designed for multiple-choice tasks. (ii) \textit{Open Evaluator:}
  Targeted at generative tasks, this evaluator supports flexible assessment protocols,
  enabling comprehensive evaluation of free-form and creative model outputs.

  \section{Experiment}

  \subsection{Benchmark Setup}

  For signal-, perception-, and semantic-level tasks, \bench\ standardizes them into a multiple-choice question format, with each question having
  four options. A correct answer is scored as 1, and an incorrect answer is scored
  as 0. Generation-level tasks usually do not have fixed-form answers. For
  Molecule-to-Spectrum tasks, the input is a molecule, and the output is a spectrum.
  For Spectrum-to-Molecule tasks, the input consists of multiple spectral images,
  and the output is a molecule. We aim to encourage models to generate meaningful
  reasoning trajectories rather than simply providing a final answer. This
  approach can help circumvent the issue of data leakage. Therefore, we use an
  additional MLLM to score the responses following these steps: (1) Model predictions
  that do not conform to the specified output format for a given question are assigned
  a score of zero. (2) For predictions meeting the required format, a dedicated scoring
  model evaluates the model's output against the answer, assigning a score
  normalized between 0 and 1. GPT-4o is employed as the scoring model in our
  experiment. This design standardizes the primary evaluation metric across all tasks
  in \bench\ to accuracy (\%).  
  Leveraging \lab's flexible model interface, we integrated 23 leading
  open- and closed-source MLLMs for our experiments. Further details on benchmarking
  candidates and cost analysis are provided in Appendix~\ref{sec:BenchmarkingCandidates}
  and \ref{sec:CostAnalysis}, respectively.



  \subsection{Main Findings}

  We draw several key insights from the results in Table~\ref{tab:vlm-tasks}.

  \noindent

  (1) \textbf{Task complexity reveals model capabilities and limitations.} Models
  exhibit strong foundational capabilities in basic tasks, with Signal and
  Perception tasks showing robust performance across all models. Spectrum Type Classification(TC)
  achieves an average accuracy of 89.09\%, while Impurity Peak Detection (ID)
  shows an average of 77.95\%. However, performance significantly declines in more
  complex tasks, particularly within the Generation category, which shows an average
  accuracy of only 6.41\%. Within the Generation level, there are notable
  performance differences: FP achieves an average of 20.29\%, significantly outperforming
  Inverse Problems (IP) at 3.50\% and \textit{De Novo} Generation (DnG) at 6.41\%.
  This suggests that models are more adept at forward prediction tasks (molecule-to-spectrum)
  than inverse problems. QE tasks prove particularly challenging, with an average
  of 30.65\% across all models, and many models scoring 0\% in IP and DnG tasks.
  This performance pattern reveals a clear hierarchy: models excel at basic
  pattern recognition and signal processing but struggle with advanced reasoning,
  creative generation, and complex cross-modal synthesis tasks that require
  deeper scientific understanding.

  \noindent
  (2) \textbf{Closed-source models lead overall performance with gemini-2.5-pro
  achieving best results.} Gemini-2.5-pro emerges as the top-performing model
  with an overall average accuracy of 67.81\%, securing top-2 scores in 6 out of
  14 tasks. The model demonstrates exceptional capabilities across multiple dimensions:
  it leads in Functional Group Recognition (GR) with 75.56\%, and ranks second
  in several other tasks, including Elemental Compositional Prediction(EP) and
  Forward Problems(FP). Closed-source models generally maintain a performance
  advantage. However, this gap is narrowing, models like InternVL3.5-241B(65.50\%)
  and InternS1-think (65.37\%) are approaching or even surpassing some closed-source
  counterparts.

  \noindent
  (3) \textbf{Reasoning capabilities drive generation task performance.} Doubao-1.5-Vision-Pro-Thinking
  demonstrates exceptional performance in generation tasks, achieving 66.67\% accuracy
  in Forward Problems (FP), significantly outperforming the second-best closed-source
  model (Gemini-2.5-pro at 50.00\%). This remarkable 16.67\% point advantage highlights
  the critical role of advanced reasoning capabilities in complex molecule generation
  tasks. InternS1-think also outperforms InternS1 (65.37\% vs. 59.57\%). This
  superior performance suggests that the ``thinking'' mode is essential for
  tackling sophisticated cross-modal scientific reasoning challenges.


  \subsection{Error Analysis}

  \begin{figure}[!htb]
      \centering
      \includegraphics[width=\linewidth]{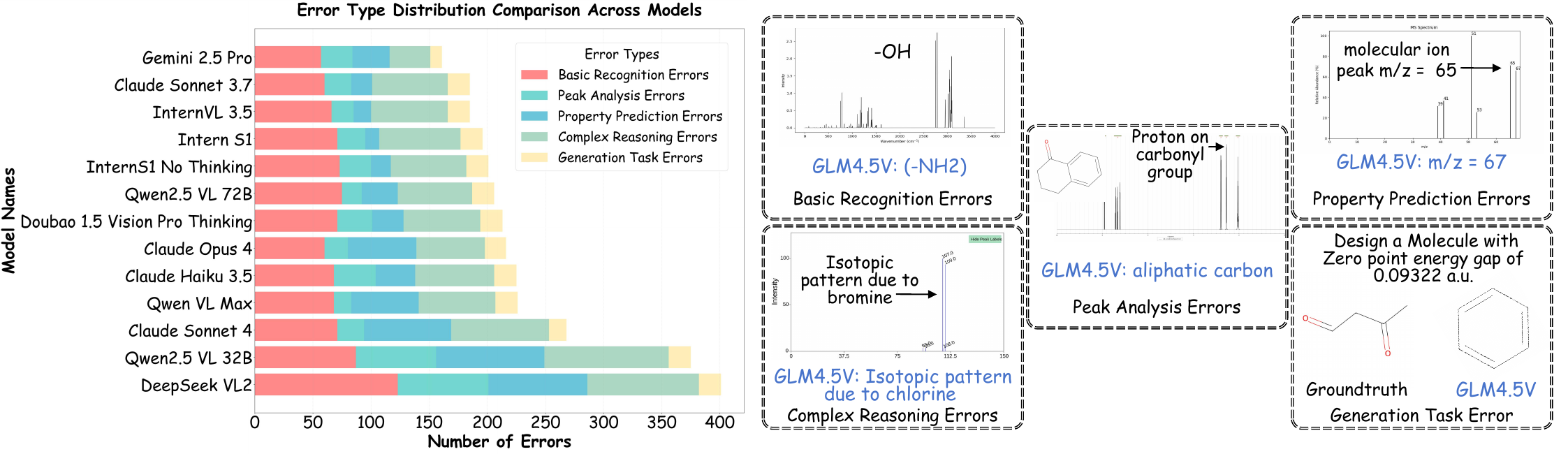}
      \caption{Error types and distributions in \bench.}
      \label{fig:error_anlysis}
  \end{figure}

\paragraph{Error Analysis.}
For our error analysis, we group the 14 tasks in \emph{\bench} into five families:
\emph{Basic recognition} (elemental composition, functional-group recognition, spectrum-type classification, spectrum-quality assessment),
\emph{Peak analysis} (peak assignment, impurity-peak detection, basic feature extraction),
\emph{Property prediction} (basic property prediction, molecular structure elucidation),
\emph{Complex reasoning} (multimodal molecular reasoning, fusing spectroscopic modalities, forward/inverse problems),
and \emph{Generation} (de novo spectrum generation).

\textbf{Family-wise minima.}
On \textbf{Basic recognition}, \textbf{Gemini-2.5-Pro} attains the lowest error (\textbf{29.1\%}),
indicating comparatively stronger grounding in spectrum type/quality and functional-group cues.
For \textbf{Peak analysis}, \textbf{Qwen2.5-VL-72B} achieves the lowest error (\textbf{14.5\%}),
suggesting effective handling of isotopic/fragment patterns and impurity peaks.
Within \textbf{Property prediction}, \textbf{Intern-S1} yields the best result (\textbf{10.5\%}), followed by
\textbf{InternVL-3.5} (\textbf{13.2\%}); both exhibit more reliable mapping from spectral evidence to molecular properties/structures.
The \textbf{Complex reasoning} slice is the sub-most challenging: although \textbf{Gemini-2.5-Pro} leads with \textbf{30.7\%} error,
the majority of models exceed $50\%$ in this family, underscoring difficulties with long-horizon, cross-modal deduction.
For \textbf{Generation}, \textbf{Gemini-2.5-Pro} again performs best (\textbf{52.6\%} error),
while several models approach failure on nearly all instances (errors near $100\%$).

\textbf{Observations and implications.}
The error profiles reveal two principal bottlenecks: (i) \emph{low-level spectral grounding}
(spectrum type/quality and functional-group perception) and
(ii) \emph{multi-step symbolic integration} across modalities and tasks.
The former dominates early-stage perception failures that cascade to peak interpretation,
whereas the latter manifests as brittle chains when executing forward/inverse reasoning or modality fusion.
We hypothesize that tighter coupling to spectroscopic priors (fragmentation and isotopic rules, impurity models)
and reasoning-aware supervision (tool-augmented peak$\rightarrow$property mappings, intermediate targets)
are necessary to reduce both recognition errors and brittle deduction.

  \section{Conclusion}
  In this work, we have presented two key contributions to advance machine learning
  in spectroscopy: \bench\ and \lab. \bench\ is a comprehensive, extensible
  benchmark suite covering over 10 spectrum modalities and 14 tasks, grounded in
  real-world chemical practices, enabling rigorous and reproducible evaluation
  across hierarchical taxonomy (signal, perception, semantic, generation). \lab\ is
  a unified, modular platform for dataset management, annotation, evaluation,
  and public leaderboards, offering a robust Python ecosystem with standardized interfaces
  that significantly lower the barrier for developing and deploying advanced models.
  Together, \bench\ and \lab\ set a new standard for spectroscopic machine learning,
  fostering systematic comparison, reproducibility, and innovation, and catalyzing
  future research for more powerful and interpretable models.

  \bibliography{iclr2026_conference}
  \bibliographystyle{iclr2026_conference}

  \newpage
 
  \appendix

\section{Ablation Study}
  \paragraph{Temperature}
\begin{table*}[!htb]
  \centering
  \small
  \setlength{\tabcolsep}{3.5pt} 
  \begin{adjustbox}{max width=\textwidth}
    \begin{tabular}{c *{12}{c}}
      \toprule
      \multirow{2}{*}{Temp} 
        & \multicolumn{4}{c}{\textbf{Qwen2.5 VL-32B}}
        & \multicolumn{4}{c}{\textbf{Qwen2.5 VL-72B}}
        & \multicolumn{4}{c}{\textbf{InternVL3-78B}} \\
      \cmidrule(lr){2-5} \cmidrule(lr){6-9} \cmidrule(lr){10-13}
        & Signal & Perception & Semantic & Generation 
        & Signal & Perception & Semantic & Generation
        & Signal & Perception & Semantic & Generation \\
      \midrule
      1.0 & 57.02 & 48.89 & 17.03 & 6.67  & 78.00 & 72.49 & 78.35 & 13.51 & 69.18 & 73.33 & 71.84 & 9.53 \\
      0.5 & 63.40 & 68.63 & 29.49 & 36.23 & 66.49 & 65.36 & 58.97 & 17.39 & 66.49 & 65.36 & 58.97 & 17.39 \\
      0.0 & 63.92 & 70.59 & 28.85 & 37.68 & 63.92 & 60.78 & 66.03 & 17.39 & 63.92 & 60.78 & 66.03 & 17.39 \\
      \bottomrule
    \end{tabular}
  \end{adjustbox}
  \caption{Performance of three models under varying temperature settings.(Top-$p$ fixed to 1)}
  \label{tab:ablation-temp}
\end{table*}

Table~\ref{tab:ablation-temp} presents the impact of varying temperature settings on three models: Qwen2.5-VL-32B, Qwen2.5-VL-72B, and InternVL3-78B.

For Qwen2.5-VL-32B, lower temperatures ($T=0.5$ and $T=0$) yield substantial improvements over $T=1.0$, particularly on the Perception, Semantic, and Generation levels. A similar trend is observed for InternVL3-78B, where deterministic decoding ($T=0$ or $T=0.5$) leads to a more balanced performance profile compared to the stochastic setting. In contrast, Qwen2.5-VL-72B behaves differently: while it achieves the highest Signal and Semantic scores at $T=1.0$, its Generation accuracy remains relatively low across all settings.

These observations indicate that smaller models tend to benefit from reduced sampling variability, as lower temperatures enhance stability and reliability. Conversely, larger models may require higher temperatures to fully exploit their expressive capacity, though this comes at the cost of weaker generative consistency.

\begin{table*}[!htb]
  \centering
  \small
  \setlength{\tabcolsep}{3.5pt} 
  \begin{adjustbox}{max width=\textwidth}
    \begin{tabular}{c *{12}{c}}
      \toprule
      \multirow{2}{*}{Top-$p$} 
        & \multicolumn{4}{c}{\textbf{Qwen2.5 VL-32B}}
        & \multicolumn{4}{c}{\textbf{Qwen2.5 VL-72B}}
        & \multicolumn{4}{c}{\textbf{InternVL3-78B}} \\
      \cmidrule(lr){2-5} \cmidrule(lr){6-9} \cmidrule(lr){10-13}
        & Signal & Perception & Semantic & Generation
        & Signal & Perception & Semantic & Generation
        & Signal & Perception & Semantic & Generation \\
      \midrule
      1.0 & 57.02 & 48.89 & 17.03 & 6.67  & 78.00 & 72.49 & 78.35 & 13.51 & 69.18 & 73.33 & 71.84 & 9.53 \\
      0.5 & 63.92 & 69.93 & 29.49 & 39.13 & 62.89 & 62.09 & 65.38 & 13.04 & 70.62 & 68.63 & 72.44 & 5.80 \\
      0.1 & 63.40 & 66.67 & 30.77 & 44.93 & 66.49 & 54.25 & 64.74 & 23.19 & 69.07 & 69.93 & 72.44 & 11.59 \\
      \bottomrule
    \end{tabular}
  \end{adjustbox}
  \caption{Performance of three models under varying Top-$p$ settings.(temperature fixed to 1).}
  \label{tab:ablation-topp}
\end{table*}

\paragraph{Top-$p$}
We further investigate the role of nucleus sampling while fixing the temperature to 1.0. The results in Table~\ref{tab:ablation-topp} show heterogeneous effects across models.

For Qwen2.5-VL-32B, reducing Top-$p$ from 1.0 to 0.1 consistently improves Semantic and Generation scores, suggesting that constraining the sampling space mitigates low-quality outputs and enhances reliability. By contrast, Qwen2.5-VL-72B attains its best Signal and Semantic results at $p=1.0$, but its Generation score is substantially reduced. Interestingly, setting $p=0.1$ recovers part of this loss, implying a trade-off between precision and diversity.

For InternVL3-78B, performance remains comparatively stable across Top-$p$ values, with minor fluctuations in Generation accuracy. This stability suggests that larger-scale models are less sensitive to sampling truncation, reflecting stronger intrinsic consistency.

\paragraph{Scoring Model}

The automatic evaluation at the Generation level uses a standardized prompt to guide the scoring models. This prompt, shown in the box below, instructs the evaluator to rate model answers on a scale from 0 to 1 based on specific rules.

\begin{promptboxblue}{Prompt Templates for OpenEvaluator}


"You are an expert evaluator. Given the following question, reference answer, and model answer, please rate the model answer on a scale of 0 to 1, and explain your reasoning.",


\textbf{Scoring rules:}
\begin{itemize}
    \item If the reference answer is an image but the model output does not contain an image, score 0.
    \item If the reference answer is text but the model output does not contain text, score 0.
    \item Otherwise, score based on the similarity and correctness of the model output compared to the reference answer.
    \item If both text and image are present, consider both in your evaluation.
\end{itemize}


``Please output your score in the format: \textbf{\textbackslash score\{X\}}, where X is a number between 0 and 1.'',


\textbf{Question:} f``Question: \{question\}''

\end{promptboxblue}

\begin{table}[t]
\centering
\small
\begin{tabular}{lccc}
\toprule
Scoring Model & FP & IP & DnG \\
\midrule
GPT-4o        & 23.33 & 0.00 & 5.26 \\
InternVL3-78B & 23.33 & 5.26 & 5.00 \\
\bottomrule
\end{tabular}
\caption{Comparison of different scoring models (GPT-4o vs. InternVL3-78B) on Generation Level evaluation. FP, IP, and DnG represent different sub-categories.}
\label{tab:ablation-scoring}
\end{table}

Since the Generation level relies on automatic evaluation, we further compare different scoring models (Table~\ref{tab:ablation-scoring}). Both GPT-4o and InternVL3-78B yield consistent FP scores (23.33\%). However, discrepancies arise in the IP dimension, where GPT-4o assigns zero while InternVL3-78B detects additional errors (5.26\%). For DnG, the results are nearly identical (5.26\% vs. 5.00\%).

These findings suggest that overall evaluation trends are robust across scoring models, but fine-grained categories may be influenced by the evaluator’s internal biases. Consequently, careful selection of the scoring model is essential to ensure fairness and reliability in automatic evaluation pipelines.

  \section{\bench\ Detailed Information}
  \label{sec:BenchDetailedInfo}

  \subsection{Signal Level}
\begin{figure}[h]
    \centering
    \includegraphics[width=0.75\textwidth]{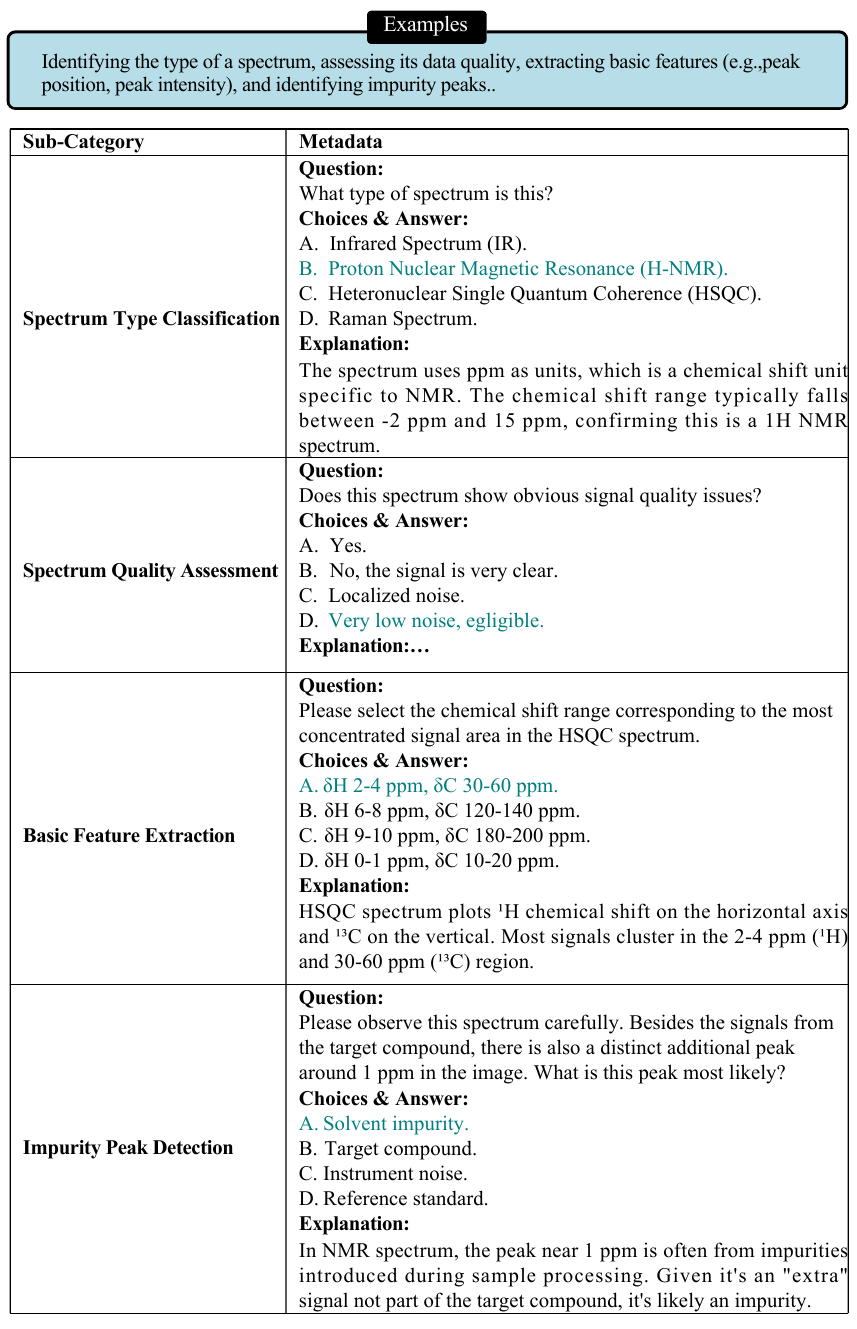}
    \caption{Example tasks and question formats at the Signal Level.}
    \label{fig:signal_level}
  \end{figure}

  This layer focuses on the direct processing and understanding of raw,
  fundamental data formats, much like extracting information from physical signals,
  as exemplified in Figure~\ref{fig:signal_level}.

  \subsection{Perception Level}
\begin{figure}[h]
    \centering
    \includegraphics[width=0.75\textwidth]{
      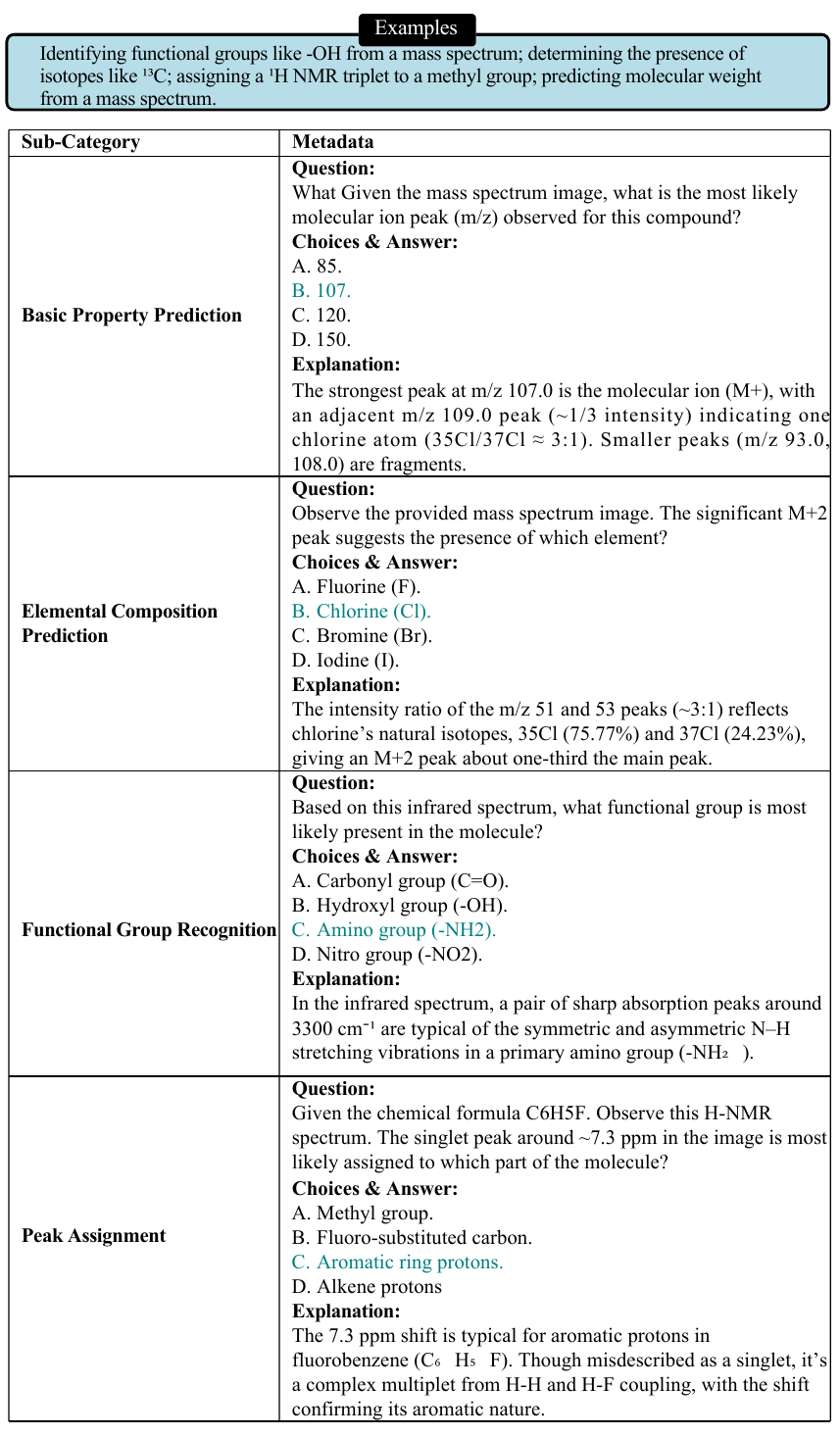
    }
    \caption{Example tasks and question formats at the perception level.}
    \label{fig:perception_level}
  \end{figure}

  This layer associates the features identified at the signal layer with
  chemical entities (functional groups, fragments, elements, and basic
  properties), as illustrated in Figure \ref{fig:perception_level}.

  \subsection{Semantic Level}
\begin{figure}[h]
    \centering
    \includegraphics[width=0.75\textwidth]{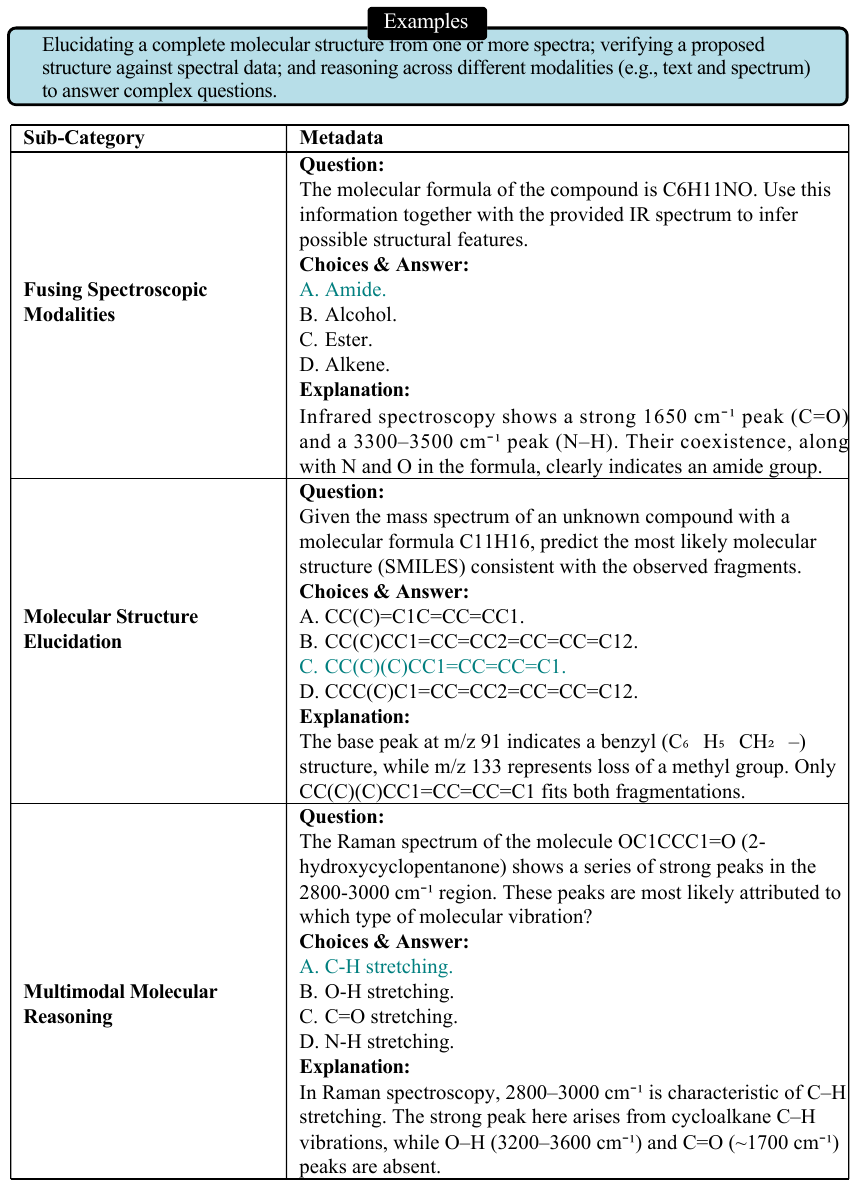}
    \caption{Example tasks and question formats at the semantic level.}
    \label{fig:semantic_level}
  \end{figure}

  This layer involves higher-level reasoning and comprehensive interpretation, connecting
  fragmented information to form complete insights or generate novel chemical structures,
  as depicted in Figure \ref{fig:semantic_level}.

  \subsection{Generation Level}

\begin{figure}[h]
    \centering
    \includegraphics[width=0.86\textwidth]{
      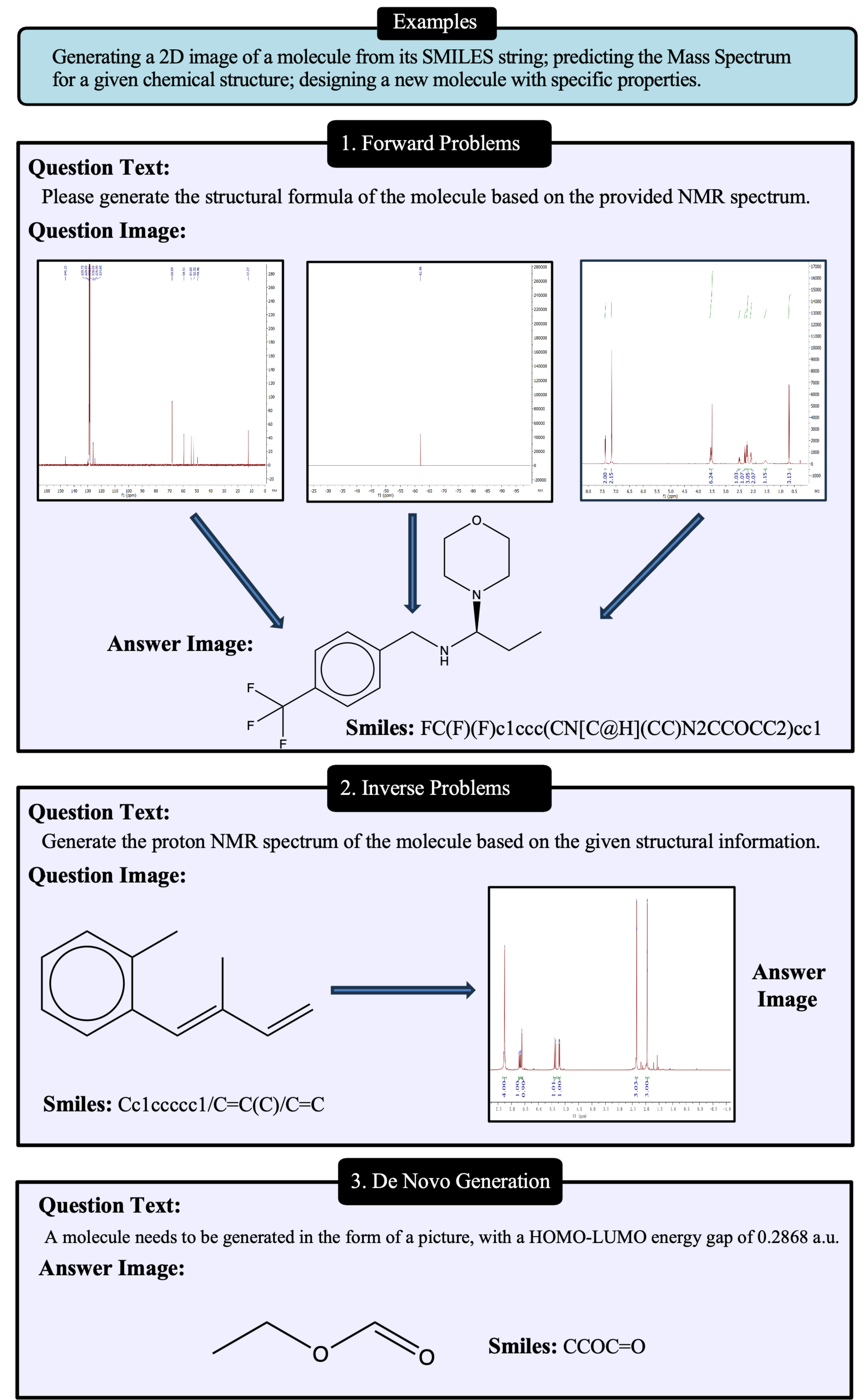
    }
    \caption{Example tasks and question formats at the Generation Level.}
    \label{fig:generation_level}
  \end{figure}

  This layer focuses on creating novel data, such as generating a 2D image of a
  molecule from its SMILES string, predicting the Mass Spectrum for a given chemical
  structure, or designing a new molecule with specific properties, as illustrated
  in Figure \ref{fig:generation_level}.

  \subsection{Data Distribution}
\begin{figure}[!htbp]
    \centering
    \begin{minipage}[t]{0.7\textwidth}
      \centering
      \includegraphics[width=\linewidth]{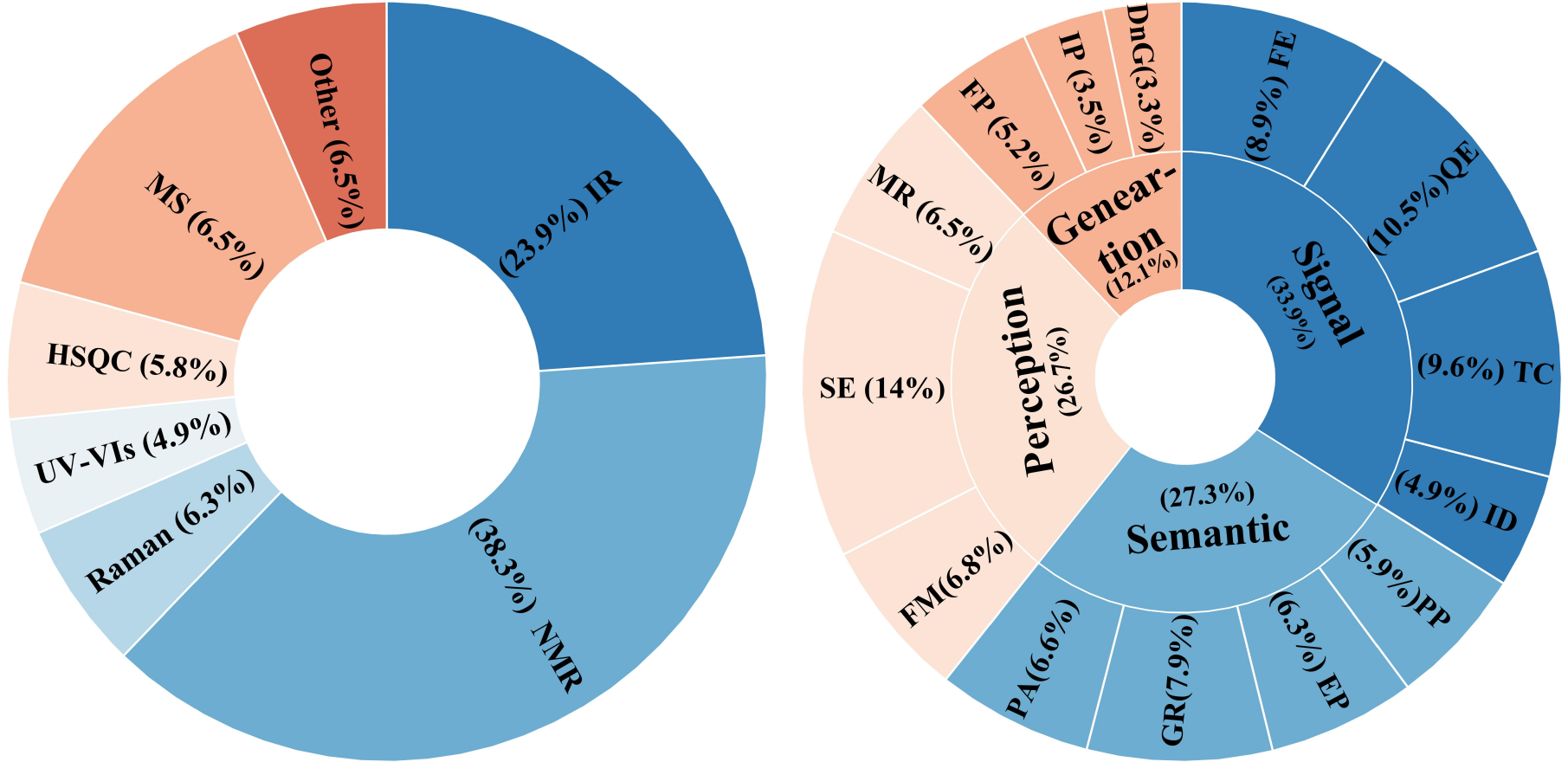}
      \caption{Distribution of spectrum types and spectroscopic task categories.
      }
      \label{fig:combined_chart}
    \end{minipage}
  \end{figure}

  To provide an overview of the data landscape, Figure \ref{fig:combined_chart}
  presents two pie charts: the left illustrates the distribution of different spectrum
  types (\textit{e.g.}, NMR, IR), while the right shows the categorization of spectroscopic
  task types. These distributions reflect the diversity of data and tasks within
  our study. It should be noted that the spectrum type statistics were generated
  by having GPT-4o scan and summarize all spectra in the benchmark. However, there
  are potential limitations: GPT may have recognition errors, and some spectrum-involving
  benchmarks lack actual image data (e.g., predicting NMR spectrum properties
  from molecular characteristics in \textit{de novo} generation tasks). Additionally, in tasks
  like multimodal fusion reasoning and forward generation problems, a single
  benchmark instance might include multiple spectra. Thus, the number of spectra
  does not align with the number of benchmarks, and this pie chart is provided only
  as a general reference.

  \section{\omnigen\ Technical Details}
  \label{sec:Annotator} In the main text, we briefly introduced the function of
  \omnigen. In this section, we will introduce its specific technical details.

  MolPuzzle~\citep{NEURIPS2024_f2b9e8e7} represents the first benchmark specifically
  designed for LLMs in spectroscopic analysis, employing a three-stage approach
  to generate question-answer pairs. While this template-based generation method
  offers efficiency, it suffers from limited coverage of spectroscopic domains
  and overly simplistic question formats. In the field of spectroscopy, high-quality
  data and benchmarks are crucial to advance AI research. The design of \omnigen\ originates
  from two key insights: First, the process of creating benchmarks shares similarities
  with the supervised data generation methods used in LLM pre-training and post-processing.
  Just as high-quality training data is essential for model performance, well-designed
  benchmarks are equally critical for evaluating and advancing the field. Second,
  we aim to utilize LLMs' few-shot and zero-shot capabilities to generate
  diverse benchmarks, enabling batch processing of seed datasets to construct large-scale
  pre-training and post-processing data. Additionally, we leverage LLMs'
  discriminative abilities for preliminary data screening and establish closed-loop
  mechanisms for continuous improvement.

  \begin{figure}[htbp]
    \centering
    \includegraphics[width=0.45\textwidth]{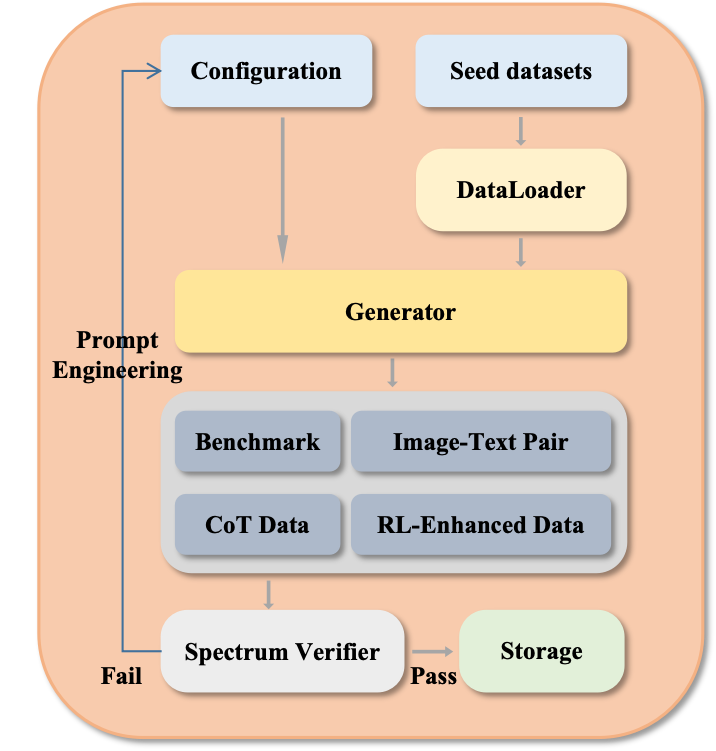}
    \caption{Technical architecture of \omnigen, illustrating the data flow from
    seed datasets through generation to quality verification.}
    \label{fig:annotator}
  \end{figure}

\begin{figure}[htbp]
    \centering
    \includegraphics[width=0.85\textwidth]{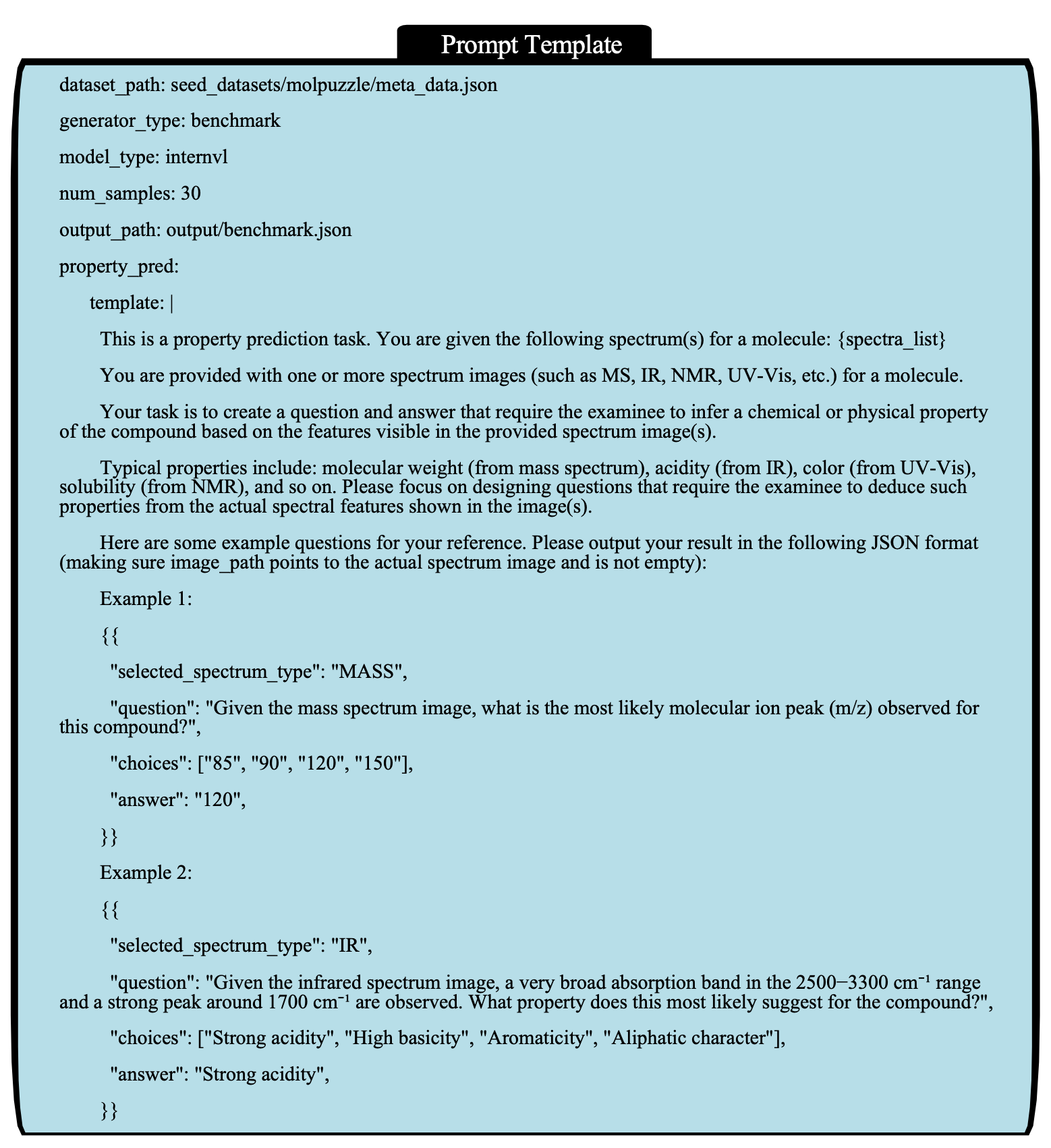}
    \caption{Example configuration for property prediction tasks, demonstrating how
    prompt templates and model parameters are specified.}
    \label{fig:configuration}
  \end{figure}

  As illustrated in Figure~\ref{fig:annotator}, \omnigen\ consists of several
  key components that work together to generate high-quality spectroscopic
  benchmarks. \textbf{Configuration \& Seed Datasets} form the foundation of the
  system. Seed datasets are extracted from multiple data sources containing essential
  spectroscopic information, while the configuration is a YAML configuration
  file that primarily configures prompt templates, instructing the generator on what
  prompts to use, along with model configurations and other parameters. As shown
  in Figure~\ref{fig:configuration}, taking property prediction as an example, the
  configuration specifies the seed datasets from MolPuzzle and provides question
  templates to guide the generator's output.

\begin{figure}[htbp]
    \centering
    \includegraphics[width=0.65\textwidth]{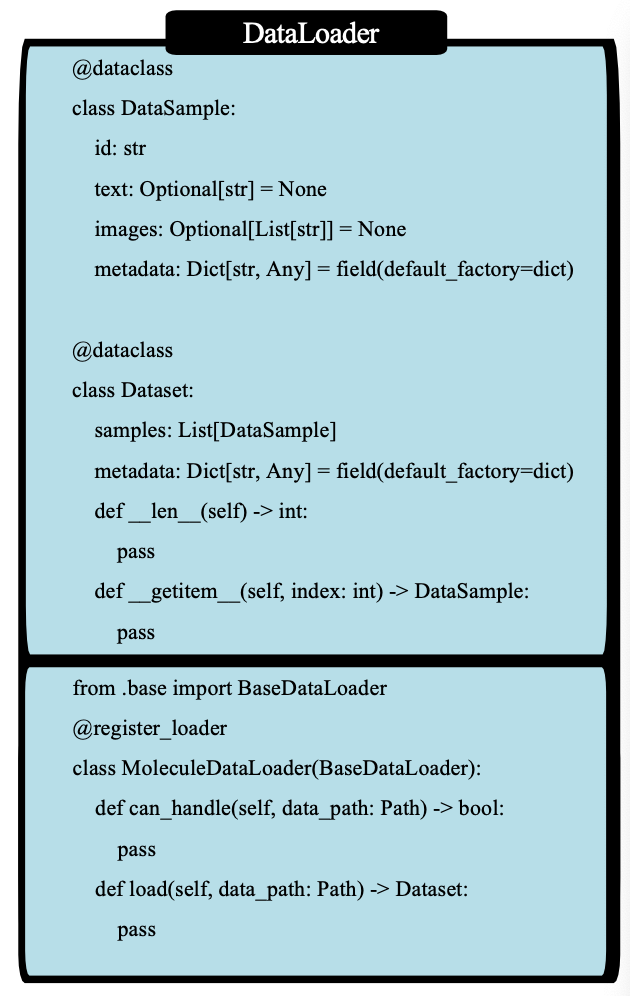}
    \caption{Plugin-based DataLoader architecture showing the registration
    mechanism for custom data loaders.}
    \label{fig:dataloader}
  \end{figure}

  \noindent
  \textbf{DataLoader} addresses the challenge of integrating diverse data sources.
  Ideally, we would like to standardize all seed datasets into a uniform format.
  However, in practice, this proves challenging as original data may possess
  complex nested file structures and diverse storage formats. To reduce adaptation
  complexity, we allow customized DataLoader designs. This design is inspired by
  PyTorch's DataLoader, which can properly load, batch, and post-process raw
  data. Our DataLoader aims to integrate various ``seed datasets'' into formats that
  can be processed by generators. The foundation consists of two base classes: DataSample,
  which represents the minimal granular information unit in \omnigen\ and serves
  as reference information for the Generator to generate individual samples; and
  Dataset, a collection of DataSample objects that provides standardized access
  methods. As demonstrated in Figure~\ref{fig:dataloader}, the DataLoader adopts
  a plugin-based architecture with an abstract registry. For different seed datasets,
  researchers only need to register their custom loaders using simple
  registration code, enabling seamless integration of diverse data sources.

  \noindent
  \textbf{Generator} operates through a three-stage workflow: First, it receives
  question templates from Configuration (including few-shot examples). Second, for
  each sample in the seed dataset, the generator uses question templates
  combined with sample metadata (such as molecular formulas, spectrum paths, SMILES
  strings, etc.) to render a prompt, which is then passed to the large language model.
  Third, the model's output is parsed into standard formats (e.g., question/choices/answer).

  \noindent
  \textbf{Quality Assurance Pipeline} ensures the reliability of generated
  benchmarks. After data generation, the system employs a multi-stage quality
  assurance process: Initial screening using rule-based methods to check data format
  and remove non-compliant samples, followed by SpectrumVerifier, a large model-based
  verification system that identifies suspicious samples requiring manual annotation.
  This closed-loop mechanism ensures that only high-quality, scientifically valid
  benchmarks are included in the final dataset. \omnigen\ will be open-sourced
  to collaborate with the research community in building a robust ecosystem and
  collectively addressing challenges in spectroscopic data generation and
  curation.

  \section{Benchmarking Candidates}

  \begin{figure}[h]
    \centering
    \includegraphics[width=\linewidth]{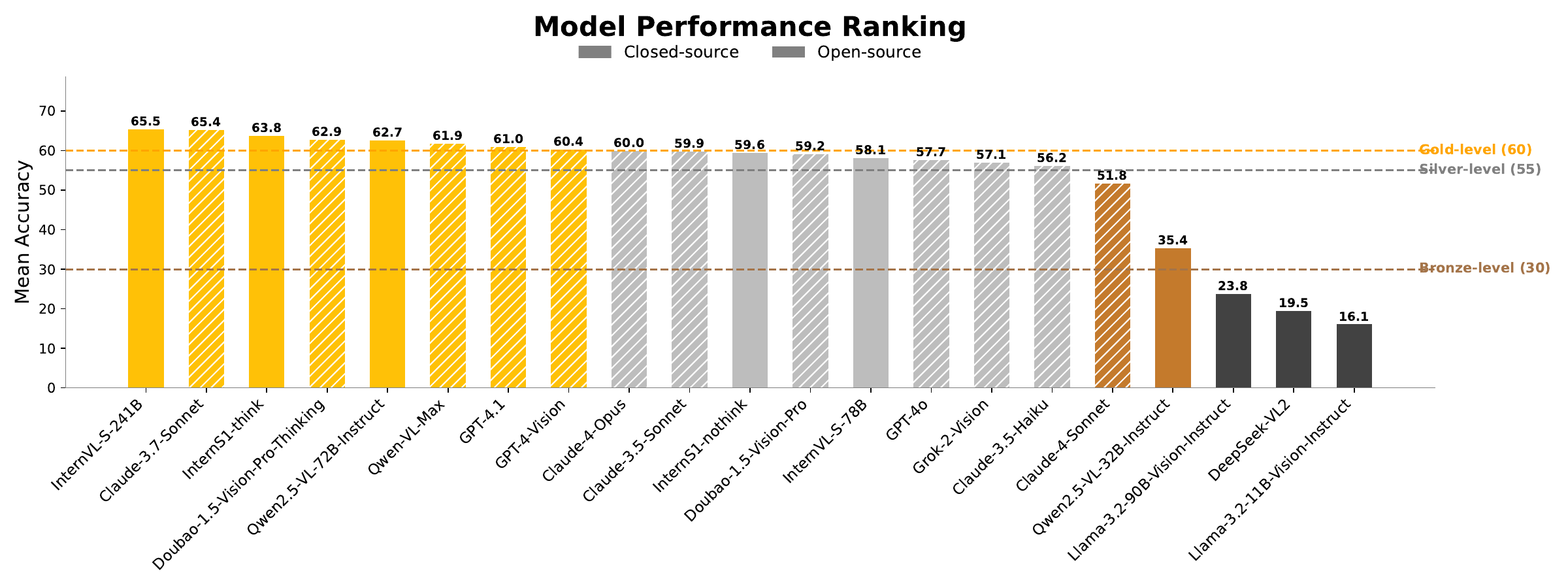}
    \caption{Performance ranking of various LLMs. }
    \label{fig:model_performance_perf1_high_res}
  \end{figure}
  
  \label{sec:BenchmarkingCandidates}
  \subsection{Open-source Models}
  \textbf{Qwen2.5-VL-32B-Instruct\citep{Qwen2.5-VL}.} Alibaba's open-source Vision-Language
  multimodal large model that handles reasoning and generation for images, text
  and video. It employs a hierarchical tagging architecture, supports multi-turn
  conversations and complex reasoning, and both the model weights and code are publicly
  available.

  \noindent
  \textbf{Qwen2.5-VL-72B-Instruct\citep{Qwen2.5-VL}.} Qwen2.5's larger-scale model
  enhances cross-modal reasoning and instruction-following capabilities,
  delivering superior performance on benchmarks such as MMMU and M3Exam while
  supporting multitasking and multilingual inputs - and is completely open-source.

  \noindent
  \textbf{InternVL3-78B \citep{chen2024expanding}.} Shanghai AI Lab releases the
  multimodal model, combining native multimodal pre-training, variable visual position
  encoding (V2PE), MPO, and test-time scaling to approach GPT-4o performance.

  \noindent
  \textbf{Llama-3.2-11b-Vision-Instruct\citep{meta2024llama3.2}.} Meta's 11 B
  lightweight multimodal model locks Llama-3.1 8 B text and pairs it with a ViT
  encoder. Two-stage training: image-text alignment then SFT+DPO, using RoPE-2D.
  Open-source.

  \noindent
  \textbf{Lllama-3.2-90b-Vision-Instruct\citep{meta2024llama3.2}.} The 90B features
  a more advanced vision adapter with cross-attention layers to inject image features
  into the LLM core. It is tuned with SFT and RLHF for enhanced performance on
  complex visual reasoning tasks.

  \noindent
  \textbf{DeepSeek-VL-2\citep{DBLP:journals/corr/abs-2412-10302}.} An open-source
  model from DeepSeek-AI featuring a Mixture-of-Experts (MoE) backbone and a dynamic
  tiling vision encoder for high-resolution images. It achieves or exceeds the
  state-of-the-art performance at the time on benchmarks like MMMU and DocVQA, with
  its code and weights fully available on GitHub.

  \noindent
  \textbf{Doubao-1.5-Vision-Pro \citep{doubao20251.5pro}.} It features a dynamic
  resolution visual encoder and MoE architecture, supporting visual QA, text-image
  matching, and image description. With billions of parameters, it shows strong generalization
  across scenarios and is available for self-hosting and fine-tuning.

  \noindent
  \textbf{Doubao-1.5-Vision-Pro-Thinking \citep{doubao20251.5pro}.} It
  integrates a ``Deep Thinking Mode'' and is trained with multi-round Reward
  Learning and reasoning style training. It excels in scientific, mathematical, and
  chain-of-thought reasoning. Supports open-source calling and API integration.

  \noindent
  \textbf{GLM-4.5V\citep{vteam2025glm45vglm41vthinkingversatilemultimodal}.} An open-source
  vision-language model from Zhipu AI and Tsinghua University that introduces a versatile
  "thinking paradigm" for enhanced reasoning. It leverages scalable
  reinforcement learning and supports full-spectrum vision reasoning, including GUI
  agent operations and code generation from screenshots.

  \noindent
  \textbf{InternS1\citep{bai2025interns1scientificmultimodalfoundation}.}A vision-language
  model developed by Shanghai AI Laboratory that features a specialized "Thinking"
  mode for enhanced multi-step reasoning. This mode allows the model to perform
  a series of self-guided logical steps to solve complex problems, particularly in
  scientific, mathematical, and logical domains.

  \subsection{Closed-source Models}
  \textbf{GPT-4o \citep{DBLP:journals/corr/abs-2410-21276}.} OpenAI's flagship ``omni''
  model natively supports text, audio, and image modalities. Delivers GPT-4-level
  intelligence with significantly faster response times and enhanced multimodal capabilities.

  \noindent
  \textbf{GPT-4.1\citep{openai2025gpt4.1}.} A reinforced version of GPT-4
  deployed through the OpenAI API, offering improved handling of complex instructions
  and logical reasoning; accepts multimodal inputs but is primarily geared
  toward text-centric tasks.

  \noindent
  \textbf{GPT-4-Vision\citep{2023GPT4VisionSC}.} A version of GPT-4 equipped with
  image input capabilities, optimized for understanding images and text and for
  the generation of conversational content, widely used for image-based Q\&A.

  \noindent
  \textbf{Claude-3.5-Haiku.} Anthropic's fastest and most cost‑effective model
  in the Claude3.5 family—offers very low latency, strong coding and reasoning
  ability, and often exceeds Claude Opus on intelligence benchmarks despite being
  lightweight.

  \noindent
  \textbf{Claude-3.5-Sonnet \citep{anthropic2024claude3.5addendum}.} Anthropic's
  multimodal large language model has mixed inference capabilities and powerful
  visual understanding functions. It supports a context of 200K tokens and is skilled
  in natural writing and code generation.

  \noindent
  \textbf{Claude-3.7-Sonnet \citep{anthropic2025claude3.7sonnet}.} An evolution
  of Claude3.5 Sonnet that introduces hybrid reasoning—users can choose between fast
  modes or step‑by‑step logical chains; offers strong task flexibility, extended
  context windows, and deep instruction‑following in multimodal settings.

  \noindent
  \textbf{Claude-4-Opus \citep{anthropic2025claude4sonnet}.} Anthropic's flagship
  model, designed for complex tasks. It boasts a powerful memory architecture and
  parallel tool invocation capabilities, and integrates with Claude Code, performing
  exceptionally in coding and reasoning benchmark tests.

  \noindent
  \textbf{Claude-4-Sonnet \citep{anthropic2025claude4sonnet}.} Claude-3.7-Sonnet's
  successor, balancing performance and speed, with low latency and high resource
  efficiency, excels in code generation.

  \noindent
  \textbf{Grok-2-Vision\citep{xai2024grok2vision}.} The multi-modal model of xAI
  combines language and visual processing capabilities to handle various images
  and documents, and supports multilingual recognition and style analysis.

  \noindent
  \textbf{Qwen-VL-Max.} The closed-source flagship model of Alibaba's Qwen
  series has been optimized for deployment in enterprise-level multimodal tasks,
  supporting joint input of images, text, videos, and others, with ultra-large parameter
  volume and high inference capability.

  \noindent
  \textbf{Gemini-2.5-Pro\citep{comanici2025gemini25pushingfrontier}.}A multimodal
  model from Google DeepMind that achieves state-of-the-art performance on frontier
  reasoning and coding benchmarks. It excels at multimodal understanding, including
  the ability to process up to 3 hours of video content and convert it into interactive
  code. Its combination of long context, multimodality, and enhanced reasoning
  capabilities unlocks new agentic workflows and complex problem-solving.

  \section{Error Cases Study}

  \subsection{Signal Level}
  We observe that the model struggles to distinguish localized noise from clean signals
  in the spectrum quality assessment task. For example, given the question ``Does
  this spectrum show obvious signal quality issues?'', the ground-truth label was
  ``Localized noise'' or ``Very low noise, eligible'', indicating minor but noticeable
  signal interference. However, the model incorrectly predicted ``No, the signal
  is very clear'', resulting in a failed case. This misclassification reveals a
  key limitation: the model tends to overestimate the clarity of the spectrum when
  the noise is not global or strongly pronounced. In visual inspection, localized
  artifacts—though subtle—can be clearly identified by human annotators, whereas
  the model often dismisses them as negligible. It lacks sufficient sensitivity to
  weak or local signal distortions, or has overfit to globally noisy or clean
  examples during training, causing it to ignore partial imperfections. This
  insight aligns with our general observation: the model often fails to distinguish
  noise from true signal, especially when the noise is spatially sparse or
  located at the margins of the image. Such behavior may stem from the fact that
  the model treats the entire spectrum as a holistic input, and lacks mechanisms
  to perform fine-grained regional quality assessment. Additionally, for models
  not inherently multi-modal, spectra are often encoded as image representations
  and then passed through vision encoders or captioning modules, potentially
  discarding low-level noise patterns. As a result, noise may not be retained in
  the model’s internal representation, leading to overly optimistic predictions.

  \subsection{Perception Level}
  \begin{figure}[!htb]
    \centering
    \includegraphics[width=0.8\linewidth]{
      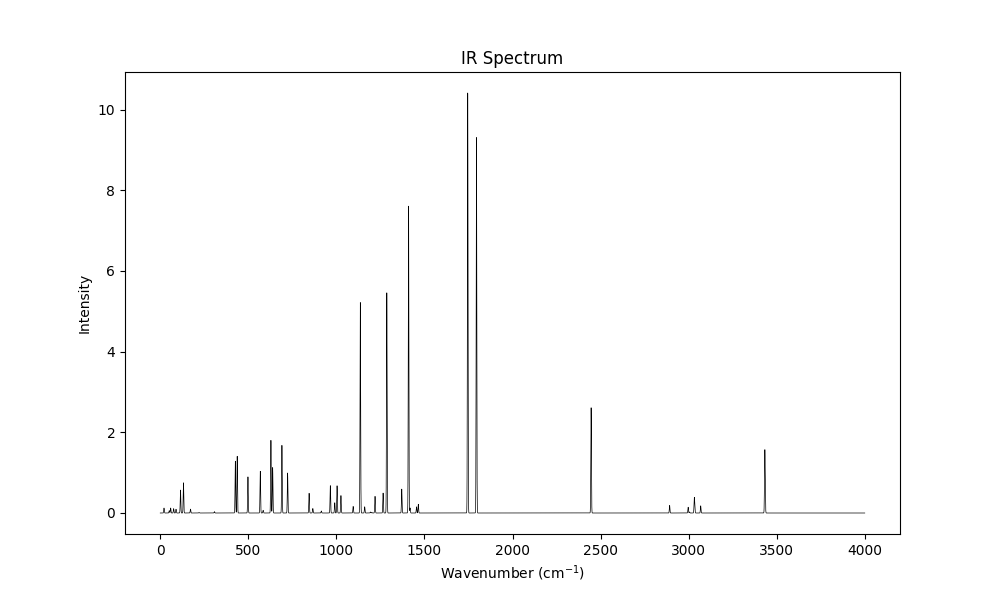
    }
    \caption{A Case of Functional Group Recognition}
    \label{fig:fgr1}
  \end{figure}
  \begin{figure}[!htb]
    \centering
    \includegraphics[width=0.99\linewidth]{
      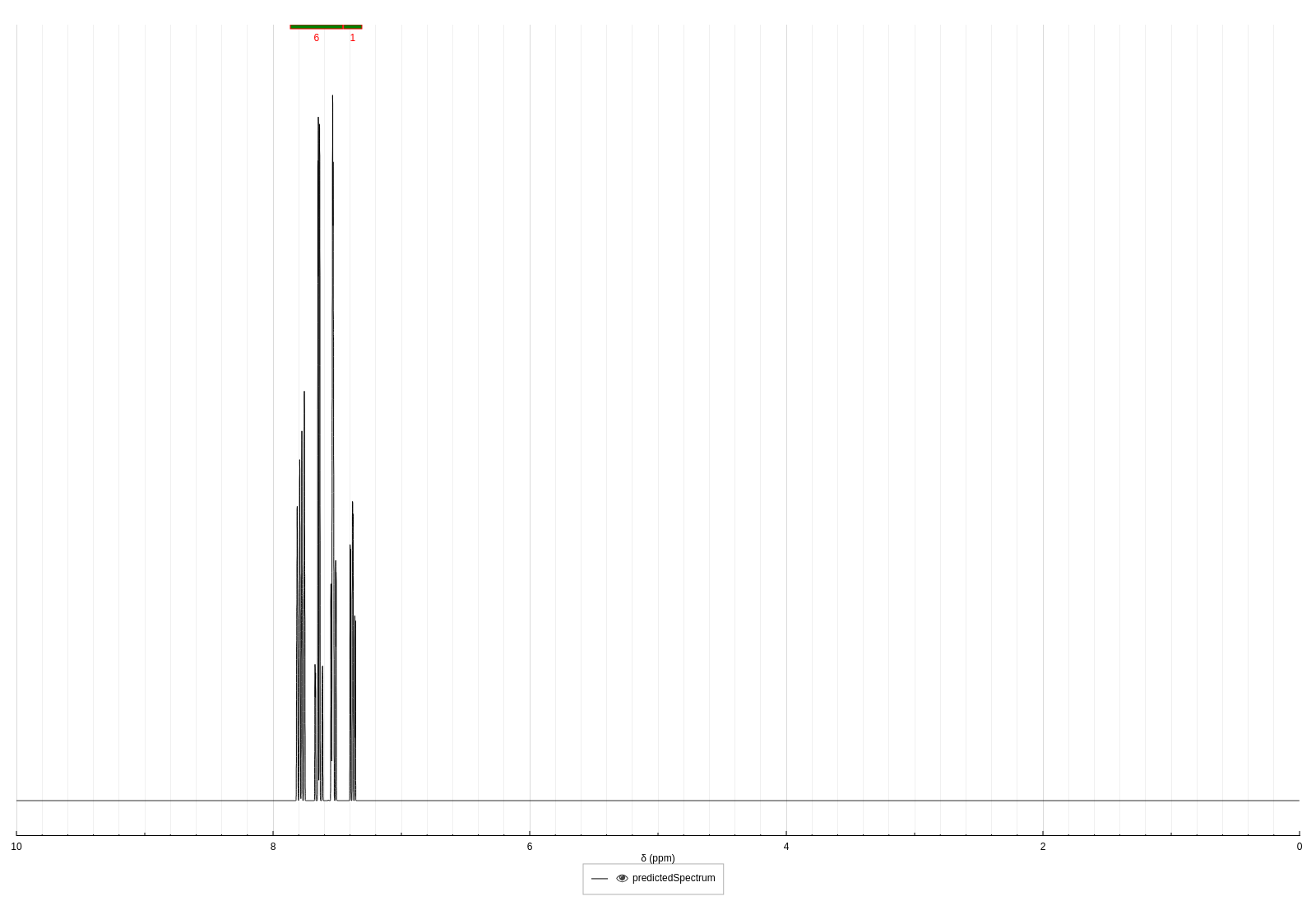
    }
    \caption{A Case of Peak Assignment}
    \label{fig:fgr2}
  \end{figure}
  We found that for functional group recognition and peak assignment tasks,
  large language models such as Doubao-1.5-pro-thinking often fail to produce
  chemically accurate predictions, even when the visual features in the spectra are
  clear to human experts. For instance, in the functional group recognition task
  (Figure \ref{fig:fgr1}), the infrared (IR) spectrum exhibits a strong
  absorption band characteristic of a \textbf{carbonyl group (C=O)}, typically
  near 1700 cm\textsuperscript{-1}. However, the model incorrectly predicted
  \textbf{hydroxyl group (-OH)}. This suggests that the model likely over-relied
  on the presence of a broad peak or baseline shift, possibly mistaking low-intensity
  or overlapping signals for OH-stretching vibrations. In the peak assignment
  task (Figure \ref{fig:fgr2}), given the molecular formula C\textsubscript{10}H\textsubscript{7}Cl
  and a clear singlet near 6.8 ppm in the \textsuperscript{1}H-NMR spectrum, the
  expected answer was \textbf{aromatic CH next to a double bond}, i.e., a non-substituted
  position in the naphthalene ring. Yet the model responded with \textbf{aromatic
  CH adjacent to Cl}, a chemically invalid assignment considering the splitting pattern
  and electronic environment. This indicates a lack of fine-grained chemical
  reasoning and possibly an overemphasis on token-level keyword association
  rather than structural context. These cases expose the model’s semantic-level misunderstanding,
  which goes beyond visual misinterpretation and highlights a deficiency in
  chemically grounded reasoning. We hypothesize two contributing factors.
  Firstly, the model may rely heavily on language priors, rather than truly
  integrating spectral visual features with molecular structure. Secondly, it
  lacks domain-specific supervision. Pretraining on generic data may not sufficiently
  expose the model to physical rules of spectroscopy, such as electron-withdrawing
  effects, chemical shift theory, or group frequency ranges.

  \subsection{Semantic Level}

  At the semantic level, tasks involving \textbf{molecular structure elucidation}
  and \textbf{multi-modal reasoning} remain particularly challenging. Consider the
  example below:

  \begin{figure}[!htb]
    \centering
    \includegraphics[width=0.8\linewidth]{
      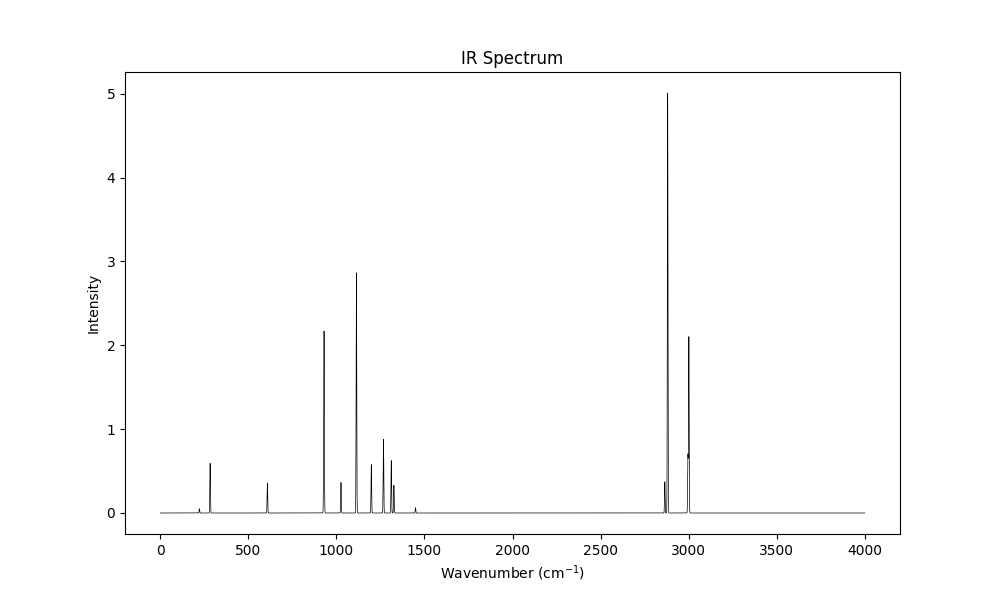
    }
    \caption{A Case of Fusing Multi-Modalities}
    \label{fig:FM1}
  \end{figure}

  In this case, the model is asked: \textit{``The molecular formula of the
  compound is C\textsubscript{4}H\textsubscript{8}O\textsubscript{2}. Use this information
  together with the provided IR spectrum image to infer possible structural features.''}
  The correct answer should be \textbf{Ether}, based on the absence of a strong carbonyl
  absorption near 1700 cm\textsuperscript{-1} and the elemental composition.
  However, the model incorrectly predicts \textbf{Carboxylic acid}, likely due to
  over-reliance on superficial signal patterns that resemble O–H stretching or C=O
  bands.

  Even when the molecular formula is omitted (pure spectrum-based reasoning),
  the model continues to produce incorrect predictions, revealing a deficiency in
  cross-modal semantic alignment. This suggests that while LLMs may perform well
  on shallow text-image associations, they struggle with integrating spectral data
  and chemical constraints in a chemically meaningful way.

  \subsection{Generation Level}
  Not surprisingly, the performance on generation tasks—especially structure
  generation—is significantly worse. This suggests that while models like \textbf{Claude-3.7-Sonnet}
  perform well on earlier levels such as perception, syntactic understanding,
  and basic semantic reasoning, they still struggle with more complex \textbf{forward
  problems} that require inferring new molecular structures from spectral data. \textbf{De
  novo generation} and \textbf{inverse problems} (e.g., predicting spectra from
  structure) pose even greater challenges, as they demand deeper chemical
  understanding and cross-modal generalization. In these settings, most models
  exhibit clear signs of overfitting or default to high-frequency patterns seen
  in training data.

  Surprisingly, \textbf{Doubao-1.5-Vision-Pro-Thinking} demonstrates promising
  performance on forward problems, aligning well with its strong results in earlier
  semantic-level tasks such as functional group recognition, peak assignment, and
  molecular structure elucidation. This consistency suggests that the model may
  have a better internal representation of cross-modal chemical semantics, though
  its capability still falls short in full generation settings.

  \section{Model Accuracy vs. Token Assumptions}
\begin{figure}[!htb]
    \centering
    \includegraphics[width=1.0\linewidth]{
      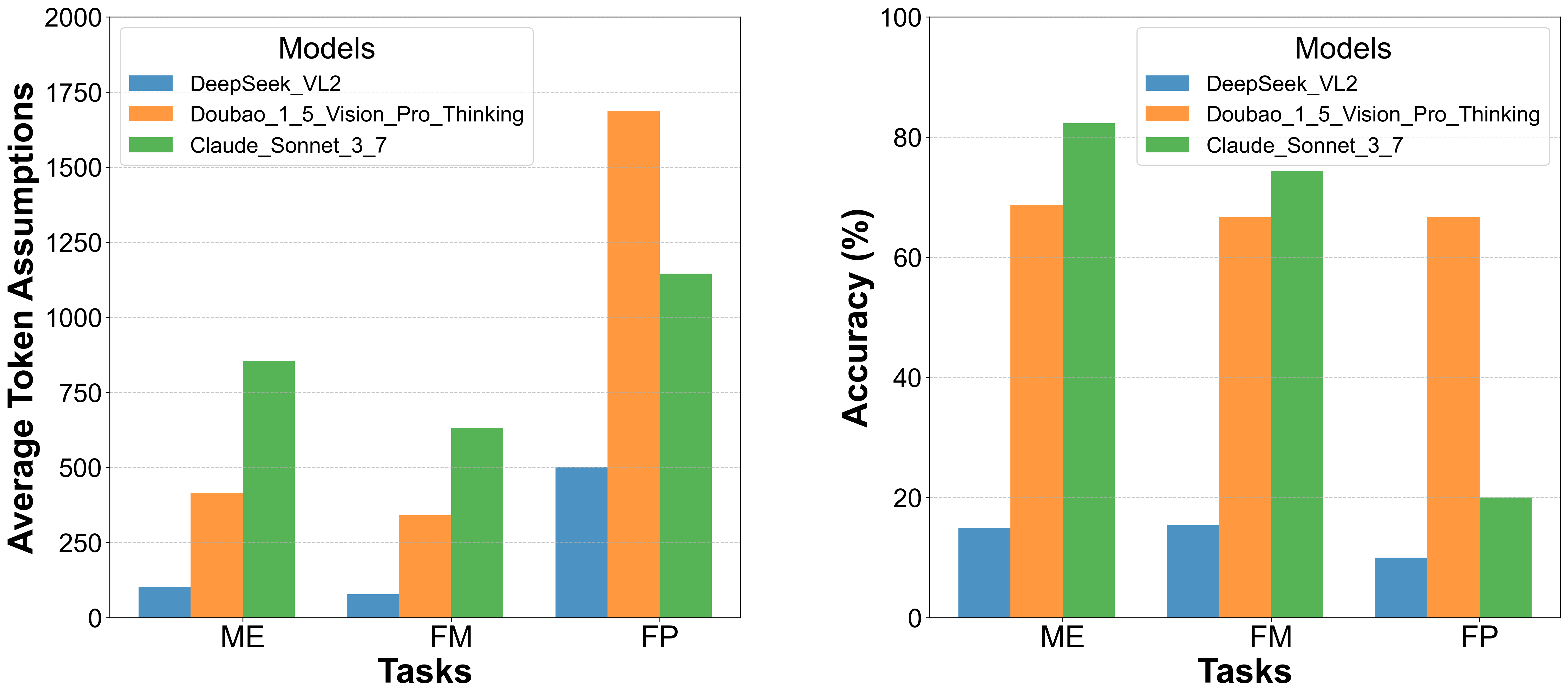
    }
    \caption{Model accuracy aligns with the model size.}
    \label{fig:token}
  \end{figure}

  We conduct a comparative analysis of several Multimodal Large Language Models (MLLMs)
  from both semantic and generative levels, focusing on three representative
  tasks: Molecule Elucidation (ME), Fusing Spectroscopic Modalities (FM), and Forward
  Problems (FP). As shown in Figure~\ref{fig:token}, the performance gap among models
  is significant. Notably, models with lower average token assumptions, such as
  \textit{DeepSeek-VL2}, tend to exhibit lower accuracy. In contrast, models
  with higher token assumptions, such as \textit{Doubao-1.5-Vision-Pro-Thinking},
  achieve superior performance, especially on complex \textit{de novo} generation
  tasks like FP. This suggests that a longer reasoning chain, reflected in higher
  token usage, benefits complex problem-solving. However, the trade-off is
  increased computational cost and significantly longer inference time. These results
  highlight the efficiency-performance dilemma in MLLMs.

  \section{Detailed Data Structure}
  \label{sec:DataStructure}
  
  This section details the comprehensive seed datasets curation pipeline and the three primary data structures that underpin our framework: the foundational \textbf{seed datasets}, the structured \textbf{benchmark data}, and the standardized \textbf{evaluation results}.

  \subsection{Seed Data Curation Details}
  The seed datasets are curated from three primary sources to ensure both diversity and scientific rigor:

\begin{enumerate}
\item \textbf{Proprietary collections and in-house experimental data}: These include unpublished spectroscopic measurements and curated datasets from our collaborating laboratories. This source comprises approximately 238,869 molecular data points covering 8 types of spectra, offering higher authenticity and usability compared to most computationally generated spectra.
\item \textbf{Public repositories and benchmark datasets}: We integrate data from a range of widely recognized and authoritative sources, including SDBS~\citep{SDBS}, QM9S~\citep{zou2023deep}, NovoBench~\citep{DBLP:journals/corr/abs-2406-11906}, and MolPuzzle~\citep{NEURIPS2024_f2b9e8e7}, among others. In total, seven distinct repositories and public datasets are used, collectively encompassing over 1.01 million unique chemical compounds.
\item \textbf{Literature mining}: Spectral data are systematically extracted from the \textit{Supporting Information} sections of peer-reviewed publications, with a focus on articles from leading journals such as the \textit{Journal of the American Chemical Society} (\textbf{JACS}) and \textit{ACS Catalysis}.
\end{enumerate}

All collected datasets undergo a unified processing pipeline that systematically maps each entry into three core chemical spaces: SMILES, molecular formula, and spectra. The resulting seed datasets are organized at the level of individual chemical substances, with each record containing the compound's SMILES, molecular formula, and a structured set of associated spectra, all stored in a standardized JSON format. This robust foundation facilitates downstream annotation and interoperability.

  \subsection{Seed Datasets Structure}
  The seed dataset is constructed by extracting essential information from raw experimental
  data, serving as the foundation for benchmark generation. Each entry contains a
  molecular index, SMILES string, molecular formula, and a list of associated
  spectra. An illustrative structure is provided in Listing~\ref{lst:seed-dataset}.
  The \texttt{path} field is a list that may contain multiple files for a given spectrum
  type, accommodating cases such as multiple mass spectra for a single molecule.


  \begin{lstlisting}[caption={Example structure of a seed dataset entry.},label={lst:seed-dataset},basicstyle=\ttfamily\footnotesize]
  {
    "molecule_index": "MOL_0001",
    "smiles": "CCCCC1=CC=CC=C1",
    "formula": "C10H14",
    "spectra": [
      {"spectrum_type": "IR", "path": ["IR/MOL_0001.png"]},
      {"spectrum_type": "MASS", "path": ["MASS/MOL_0001.jpg", 
      "MASS/MOL_0001_2.jpg"]},
      {"spectrum_type": "C-NMR", "path": ["C-NMR/MOL_0001.png"]},
      {"spectrum_type": "H-NMR", "path": ["H-NMR/MOL_0001.png"]}
    ]
  }
  \end{lstlisting}

  \subsection{\bench\ Data Structure}

  The benchmark data structure is designed to support a diverse range of tasks,
  including signal interpretation, perception, and semantic understanding. Each entry
  includes a unique identifier, image path(s), question, answer choices, ground
  truth answer, category, sub-category, data source, and timestamp. A representative
  example is shown in Listing~\ref{lst:benchmark-data}. After processing by \lab,
  three additional fields are appended: \texttt{model\_response} (the model's
  reasoning and output), \texttt{model\_prediction} (the answer extracted from
  the model response), and \texttt{pass} (a boolean indicating whether the model's
  prediction matches the ground truth).

  \begin{lstlisting}[caption={Example of a benchmark data entry.},label={lst:benchmark-data}, basicstyle=\ttfamily\footnotesize]
  {
    "id": "Perception_a9cf_250723_235951_318294",
    "image_path": [
      "data/Perception/Basic Property Prediction/Perception_a9cf_q.png"
    ],
    "question": "Given the mass spectrum image, what is the most likely 
    molecular ion peak (m/z) observed for this compound?",
    "choices": ["85", "90", "120", "133"],
    "answer": "133",
    "category": "Perception",
    "sub_category": "Basic Property Prediction",
    "source": "",
    "timestamp": "2025-07-23 23:59:51"
  }
  \end{lstlisting}

  \subsection{Evaluation Results Structure}
  The evaluation results structure records the model's predictions and
  performance for each benchmark instance. Listing~\ref{lst:evaluation-results} illustrates
  the format. For all data structures, the \texttt{image\_path} field is
  specified relative to the \texttt{data} directory to ensure clarity and reproducibility.
  This standardized design facilitates systematic benchmarking and transparent
  evaluation across a wide range of spectroscopic machine learning tasks.

  \begin{lstlisting}[caption={Example of an evaluation results entry.},label={lst:evaluation-results},basicstyle=\ttfamily\footnotesize]
  {
    "id": "Signal_9131_250723_110552_245529_2",
    "image_path": [
      "data/Signal/Spectrum Type Classification/Signal_9131_2_q.png"
    ],
    "question": "What type of spectrum is shown in the image?",
    "choices": [
      "Infrared Spectrum (IR)",
      "Proton Nuclear Magnetic Resonance (H-NMR)",
      "Mass Spectrometry (MS)",
      "Carbon Nuclear Magnetic Resonance (C-NMR)"
    ],
    "answer": "Mass Spectrometry (MS)",
    "category": "Signal",
    "sub_category": "Spectrum Type Classification",
    "source": "",
    "timestamp": "2025-07-23 11:05:52",
    "model_prediction": "Mass Spectrometry (MS)",
    "model_response": "\\answer{Mass Spectrometry (MS)}",
    "pass": true
  }
  \end{lstlisting}

  \section{Cost Analysis}
  \label{sec:CostAnalysis}

  To ensure consistency and fairness across all experiments, \lab\  employs a
  unified model interface and conducts all inference via API services, regardless
  of whether the underlying models are open-source or proprietary. This
  standardized evaluation pipeline enables direct and equitable comparison of
  model performance. With the exception of the generation-level scoring model,
  each benchmark run requires an average of 572 model invocations. The use of remote
  APIs introduces network latency, resulting in variability in inference times. Depending
  on the model architecture and complexity, the total time required to complete
  the full \bench\ benchmark ranges from approximately 40 minutes to 2 hours. For
  each model, we systematically record the overall inference time and the
  estimated monetary cost associated with completing the benchmark.

  Given the current benchmark prompts and \lab's prompt engineering design, a complete
  run of the benchmark requires approximately 1,219,083 input tokens and 41,522 output
  tokens (as measured on InternVL3-78B, this figure is provided for reference only).
  Models with more elaborate reasoning or ``thinking'' capabilities may incur even
  higher token consumption.

  Table~\ref{tab:resource_cost} summarizes the key statistics for representative
  models evaluated in this study. 

  \begin{table}[h]
    \centering
    \caption{Resource consumption and cost for representative models on the full
    \bench\ benchmark.}
    \label{tab:resource_cost}
    \begin{tabular}{lccc}
      \toprule \textbf{Model}   & \textbf{Inference Time (min)} & \textbf{Cost (USD)} \\
      \midrule Claude-3.5-Haiku & 99                            & \$0.94              \\
      Claude-3.5-Sonnet         & 70                            & \$7.47              \\
      Claude-4-Opus             & 123                           & \$24.00             \\
      Claude-4-Sonnet           & 90                            & \$11.66             \\
      GPT-4o                    & 103                           & \$4.23              \\
      GPT-4-Vision-Preview      & 113                           & \$8.08              \\
      GPT-4.1-2025-04-14        & 103                           & \$1.54              \\
      Grok-2-Vision             & 62                            & \$2.12              \\
      InternVL3-78B             & 120                           & N/A                 \\
      \bottomrule
    \end{tabular}
  \end{table}

  \section{Usage of Large Language Models in This Manuscript}
  In preparing this manuscript, we used a large language model (LLM) solely for editorial purposes. Its functions were limited to proofreading for typographical errors, correcting grammatical mistakes, and enhancing the clarity and readability of the text.

  \section{Limitations}
  While this work introduces the concept of \name\ , it is important to acknowledge
  that the field of AI for Spectroscopy remains in its nascent stages, we
  recognize several limitations within our primary contributions, \bench\ and \lab\ .

  \noindent
  \textbf{Limitations of \bench} First, regarding Task Format, \bench\ currently
  supports only multiple-choice and a limited number of open-ended questions.
  While this design is suitable for Large Language Models (LLMs), it is insufficient
  for evaluating a broader range of machine learning models, such as
  Convolutional Neural Networks (CNNs) and Graph Neural Networks (GNNs), as discussed
  in our introduction. Second, concerning Spectrum Type, although we have
  incorporated a wide array of spectrum types compared to previous works~\citep{lu_vib2mol_2025, Xu2025-qr,DBLP:conf/nips/BushuievBJYKSHW24,DBLP:journals/corr/abs-2406-11906},
  several crucial spectroscopic modalities remain uncovered. Notable examples include
  X-ray Diffraction (XRD) \citep{Guo2024,Salgado2023} and fluorescence spectra
  \citep{parker_fluorescence_1962}, which are vital for comprehensive material
  characterization. Finally, addressing Spectroscopic Task Type, spectroscopy techniques
  are fundamental across diverse scientific disciplines, including physics, astronomy,
  chemistry, and biology, primarily for characterizing substances like molecules,
  proteins, peptides, and SMILES sequences. From the perspective of LLMs, a generic
  categorization of modalities into ``text'' and ``images'' is inadequate for representing
  the complexity of data. The inherent diversity of spectroscopic modalities
  complicates the immediate definition of all possible tasks. Consequently,
  \bench\ presently lacks important benchmarks in several areas, such as spectrum-spectrum
  retrieval \citep{Curry1969, Wang2022, lu_vib2mol_2025} and peptide sequence
  analysis \citep{DBLP:journals/corr/abs-2406-11906}. We acknowledge that it
  will be challenging for \bench\ to encompass all relevant tasks in the near
  future, and we aim to foster collaborative efforts with the community and various
  laboratories to collectively advance the development of AI in spectroscopy.

  \noindent
  \textbf{Limitations of \lab} Our second main contribution, \lab, also presents
  certain limitations. Firstly, regarding its data functionality, while \lab\ successfully
  unifies seed datasets and provides data curation tools-\omnigen, it currently
  lacks tools for the preprocessing and segmentation of raw data across multiple
  spectroscopic modalities. Secondly, concerning metrics, the current evaluation
  framework within \lab\ is relatively simplistic, relying primarily on accuracy
  and a lenient, LLM-based scoring method for open-ended questions. In future
  iterations, we plan to define and incorporate a broader array of task-specific
  metrics to enable more nuanced and robust model evaluation.
\end{document}

%% file: math_commands.tex
\usepackage{amsmath,amsfonts,bm}

\def\eqref#1{equation~\ref{#1}}

\def\1{\bm{1}}

\DeclareMathAlphabet{\mathsfit}{\encodingdefault}{\sfdefault}{m}{sl}
\SetMathAlphabet{\mathsfit}{bold}{\encodingdefault}{\sfdefault}{bx}{n}

